\documentclass{article} % For LaTeX2e
\usepackage{iclr2026_conference,times}
\usepackage[final]{graphicx}
\usepackage{amssymb}
\usepackage{amsmath}
\usepackage{wrapfig}
\usepackage{subcaption} % Required for subfigures
\usepackage[most]{tcolorbox}

\usepackage{amsmath,amsfonts,bm}

\def\eqref#1{equation~\ref{#1}}
\def\1{\bm{1}}

\DeclareMathAlphabet{\mathsfit}{\encodingdefault}{\sfdefault}{m}{sl}
\SetMathAlphabet{\mathsfit}{bold}{\encodingdefault}{\sfdefault}{bx}{n}

\usepackage{hyperref}
\usepackage{url}
\usepackage{xcolor}

\newif\ifshowedits      % <--- define the boolean
\showeditstrue          % <--- or \showeditsfalse to hide edits

\ifshowedits
  \DeclareRobustCommand{\edit}[1]{\textcolor{black}{#1}}
\else
  \DeclareRobustCommand{\edit}[1]{#1}
\fi

\title{What We Don't C: Manifold Disentanglement for Structured Discovery}

\newcommand{\myauthorblock}{%
  \makebox[0.3\textwidth][c]{\textbf{Brian Rogers}$^{1}\thanks{%
    Corresponding author: \texttt{brian.rogers@physics.ox.ac.uk}. 
    $^{1}$Oxford Astrophysics, University of Oxford; 
    $^{2}$Ellison Institute of Technology; 
    $^{3}$Breakthrough Listen, UC Berkeley; 
    $^{4}$SETI Institute; 
    $^{5}$OxRSE, Doctoral Training Centre, University of Oxford; 
    $^{6}$Centre for Data Intensive Science, UCL. 
    }$} \hspace{\fill} 
  \makebox[0.3\textwidth][c]{\textbf{Micah Bowles}$^{1,2}$} \hspace{\fill} 
  \makebox[0.3\textwidth][c]{\textbf{Chris J. Lintott}$^{1}$} \\
  \vspace{0.3em}
  \makebox[0.3\textwidth][c]{\textbf{Steve Croft}$^{1,3,4}$} \hspace{\fill} 
  \makebox[0.3\textwidth][c]{\textbf{Oliver N. F. King}$^{5}$} \hspace{\fill} 
  \makebox[0.3\textwidth][c]{\textbf{James Kostas Ray}$^{6}$}
}
\newtcolorbox{contextbox}[1][]{%
  enhanced,
  breakable,
  colback=blue!3,
  colframe=blue!65!black,
  boxrule=0.8pt,
  arc=2mm,
  left=2mm,right=2mm,top=1.2mm,bottom=1.2mm,
  fonttitle=\normalsize,
  attach title to upper,
  #1
}

\author{\myauthorblock}

\author{\myauthorblock}

\author{\myauthorblock}

\iclrfinalcopy % Uncomment for camera-ready version, but NOT for submission.
\begin{document}

\maketitle

\begin{abstract}
Accessing information in learned representations is critical for annotation, discovery, and data filtering in disciplines where high-dimensional datasets are common. We introduce \textit{What We Don't C}, a novel approach based on latent flow matching that disentangles latent subspaces by explicitly removing information included in conditional guidance, resulting in meaningful residual representations. This allows factors of variation which have not already been captured in conditioning to become more readily available. We show how guidance in the flow path necessarily represses the information from the guiding, conditioning variables. Our results highlight this approach as a simple yet powerful mechanism for analyzing, controlling, and repurposing latent representations, providing a pathway toward using generative models to explore \textit{what we don't capture, consider, or catalog}.
\end{abstract}

\begin{contextbox}
\textbf{A Note on Recent Work.} There has been a burst of publications since this manuscript was written on representation learning for generative tasks, particularly on constraining latent variables to lie on Gaussian or spherical manifolds. For example, recent generative-centric work includes the Sphere Encoder \citep{yue2026imagegenerationsphereencoder}, Riemannian Flow Matching with Jacobi Regularization \citep{kumar2026learningmanifoldunlockingstandard} and the distribution matching VAE \citep{ye2025distributionmatchingvariationalautoencoder}. Further works include the Wristband Loss \citep{mvparakhin_ml_tidbits_wristband} and applications of InfoNCE \citep{betser2026infonce}, both of which aim to map latent manifolds to Gaussian distributions. The LeJEPA paper \citep{balestriero2025lejepa} suggests that an optimal embedding distribution for foundation models is isotropic unit Gaussians and the paper provides regularisation techniques to promote this structure.
We emphasize that WWDC is similar to these works in that it makes use of the geometry of embeddings for better sampling and information retrieval. We especially note that the Wristband Loss addresses some similar challenges but does not take into account that flow matching (including single-step variants such as mean flow matching) preserves structure through optimal transport constraints. This is the key insight on which WWDC relies to enable controlled generative modeling, manifold disentanglement, and geometric control of the latent space.
\end{contextbox}

\section{Introduction}
% Motivation
Representation learning aims to map a dataset to a lower-dimensional data manifold where each data point is represented by a lower-dimensional vector on this manifold. These representations have found many uses including data filtering \citep{Liang2018CollaborativeFiltering}, searching \citep{khattab2020colbert}, clustering \citep{ren2024deep}, labeling \edit{\citep{YM.2020Self-labelling}}, outlier detection \citep{han2022adbench}, visualization \citep{mcinnes2018umap, JMLR:v9:vandermaaten08a} \edit{and more} \citep{bengio2013representation}. Approaches to representation learning are often validated on known features of the underlying data\edit{. Any }methodological improvements are measured against benchmarks, such as how well \edit{a linear or non-linear model can recover certain features from the representations}. In practice, \edit{approaches to representation learning often remain unchanged even when supervised labels are available.}

\begin{figure}
    \centering
    \includegraphics[width=0.95\linewidth]{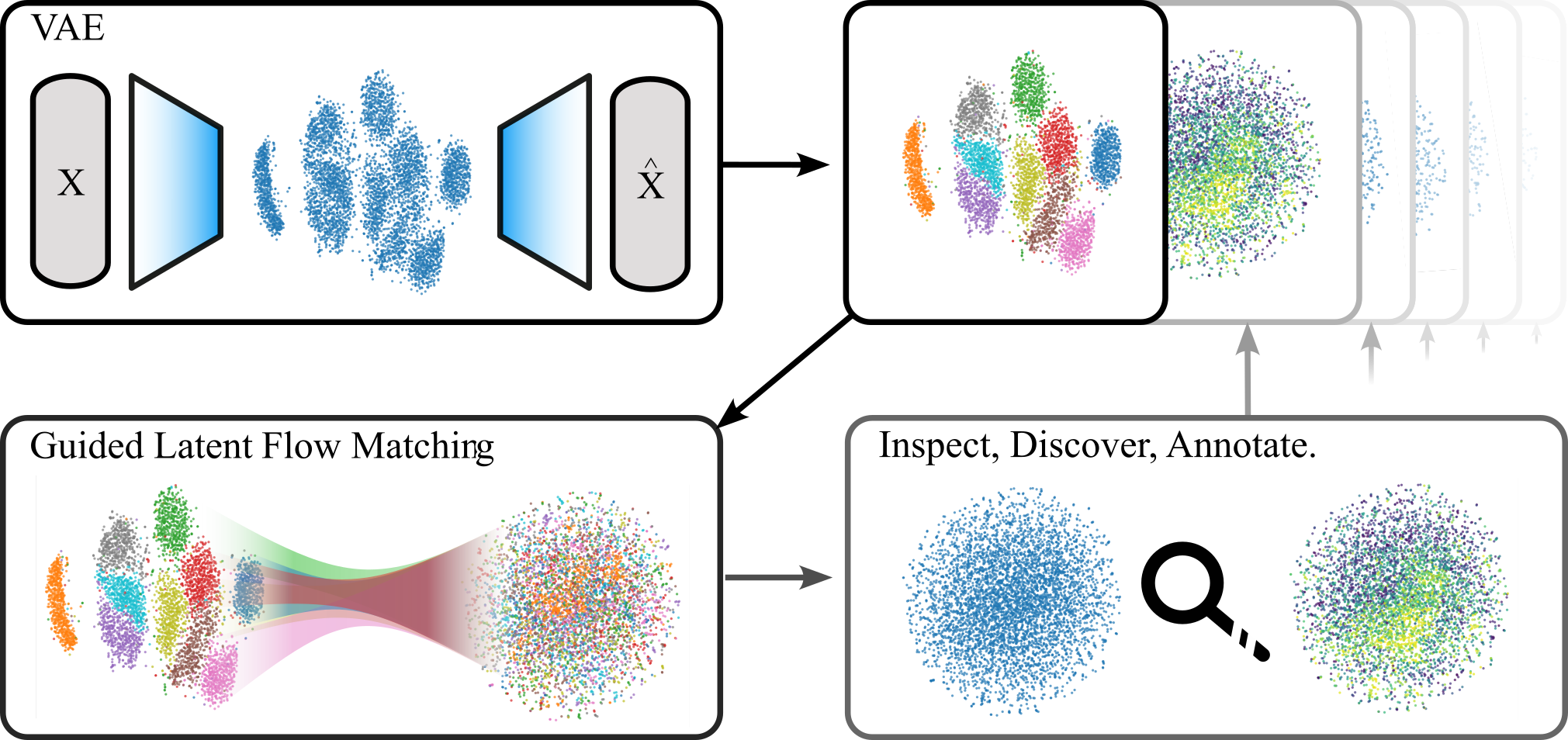}
    \caption{\edit{Scientific discovery of What We Don't C (WWDC). The representations of a VAE can be used to annotate data. These (now) known features can then be used to condition a flow, removing them from the manifold. This new disentangled manifold can then be inspected to uncover new features of the data. With sufficient coverage, these can then be used to continue the cycle, enabling access to further features.}}
    \label{fig:wwdc}
\end{figure}

% Gap
\edit{In this work, we introduce \emph{What We Don't C} (WWDC), an approach that disentangles known features from existing data manifolds. We make no requirement of our approach to fully separate all features into individual dimensions and instead aim only to separate features from a given manifold, enabling applications to real-world datasets and features.
Conditioning on known features of a dataset, we use representations \citep[e.g. from VAE;][]{kingma2014semi, JimenezRezende2014StochasticBA} to flow match \citep{lipman2023flowmatchinggenerativemodeling} to a base distribution. Far from what is often just considered a base or `noise' distribution, our work highlights that substantial structure is preserved from the representations. When flowing to the base distribution using conditioning guidance, features belonging to the conditioning are \textit{suppressed}, which allows easier access to otherwise obfuscated features.
Having identified its potential as a tool for scientific discovery, we illustrate how WWDC can be used in a discovery engine in Figure~\ref{fig:wwdc}.}

In this work we:
\begin{itemize}
    % 0. Approach
    \item Detail \edit{an approach} which uses latent guided flow matching to enable structured discovery without conflating the most dominant signals that have already been thoroughly \textit{captured}, \textit{considered}, and \textit{cataloged}, moving to uncover meaningful representations of \textit{what we don't see}.
    % 1. Theory
    \item Highlight theoretical arguments for the removal and preservation of relevant structures in the manifold. 
    % 2. Synthetic Gaussian 2d
    \item \edit{Validate our geometric understanding of conditional flows and their base distributions on a fully synthetic dataset.
    \item Verify the approach through controlled experiments on a colored variant of MNIST.}
    % 3. Real data / Real science
    \item Demonstrate the utility of this method for \edit{real-world datasets} by directly isolating the disentangled features of real galaxy images.
\end{itemize}

% Structure 
Section~\ref{sec: Background} discusses and introduces the relevant background material. Section~\ref{sec: approach} explains the approach adopted in WWDC including why the resulting space maintains useful structures from the original manifold. Section~\ref{subsec: results} presents results across increasingly complex datasets and problems. These experiments span various degrees of complexity from simple 2D Gaussians in Section~\ref{subsec: 2d gaussians}, a colored variant of MNIST in Section~\ref{subsec: cmnist}, and astrophysics galaxy morphology in Section~\ref{subsec: galaxy10}. 
Section~\ref{sec: Conclusion} summarizes the work and discusses future directions.
Comprehensive details and additional generative samples are provided in the appendices.

\section{Background}
\label{sec: Background}
\subsection{Manifold Disentanglement}
\edit{In this work, we adopt a different approach to disentanglement which we term manifold disentanglement. Using existing representations and known data features, we create new representations conditioned on the known features. This creates a rich, residual representation which we explore in Section~\ref{sec: approach} and Section~\ref{subsec: results}. Notably, manifold disentanglement requires an existing manifold to disentangle, meaning this problem is defined for frozen, and pre-trained representation learning models. This cuts computational costs significantly, which is important for the envisioned application: the iterative discovery of new signals of interest in data as outlined in Figure~\ref{fig:wwdc}.}

\edit{This methodology differs from existing notions of disentanglement in the literature. In particular, unsupervised disentanglement learning is one approach to disentanglement, which hypothesizes that realistic data can be generated by a few explanatory factors of variation and these can be learned by unsupervised algorithms \citep[]{higgins2017betavae, burgess2018understandingdisentanglingbetavae, chen2019isolatingsourcesdisentanglementvariational, jeong2019learning, shao2020controlvae, DBLP:conf/iclr/0001SB18}. However, by design, these approaches do not incorporate supervised signals into the disentanglement process; yet such approaches typically require ground-truth factors of variation for evaluation \citep{higgins2017betavae, chen2019isolatingsourcesdisentanglementvariational} making them infeasible for complex datasets where factors of variation are unknown or entangled.}

\edit{Other approaches have incorporated supervised signals when learning representations, using control of input transformations \citep{NIPS2015_ced556cd, 10.5555/3666122.3668288}, a cross-covariance penalty to encourage linear independence \citep{cheung2015discoveringhiddenfactorsvariation} or explicit priors and computational graph structures \citep[e.g.,][]{kingma2014semi, NIPS2015_8d55a249, siddharth2017learningdisentangledrepresentationssemisupervised}. These approaches are inflexible because they require either knowledge of input transformations or a specialized computational graph structure which is often infeasible due to feature complexity or computational constraints. Further, these approaches do not leverage existing representations and therefore require full retraining for newly proposed conditioning variables. Our goal is to re-purpose existing representations in a flexible way using flow matching because this allows efficient training of models for many different proposed conditioning variables and ultimately, enables the iterative process of discovery shown in Figure~\ref{fig:wwdc}.}

\edit{Given the limitations we have outlined above, we seek to re-purpose existing representations from pre-trained VAEs using guided latent flow matching. In particular, we use a classifier-free approach during training of the latent flow model. In the next section, we provide a brief introduction to each of these three components of our approach in WWDC.}

\subsection{Methods}
\label{subsec: methods}

\textbf{Variational autoencoders (VAE)}: VAEs learn latent variables $z$ from data $x$ by maximizing the following amended variational lower bound \citep{kingma2022autoencodingvariationalbayes, JimenezRezende2014StochasticBA, higgins2017betavae}

\begin{equation}
    \label{eqn:VAE}
    \mathcal{L}_{\theta, \phi, \beta} = \mathbb{E}_{q_{\phi}(z|x)} \left[ \log p_{\theta}(x|z) \right] - \beta D_{\text{KL}}(q_{\phi}(z|x) \ || \ p(z))
\end{equation}

$q_{\phi}(z|x)$ is a neural network often referred to as the encoder which is used to approximate the posterior of the latent variables $z$ given data $x$. $p_{\theta}(x|z)$ is also parameterized by a neural network, known as the decoder, which reconstructs the data from the latent variables. $ D_{\text{KL}}(q_{\phi}(z|x) || p(z))$ is the Kullback-Leibler (KL) divergence \citep{kullback1951information} between the encoder distribution and a prior on the latent variables, often chosen to be an isotropic unit-variance Gaussian. $\beta$ is a hyperparameter that weights the divergence.

\textbf{Flow matching}: Flow matching has emerged as a simple yet efficient framework for generative modeling by interpolating couplings between an arbitrary source distribution and target distribution; it boasts recent achievements in image generation \citep{esser2024scalingrectifiedflowtransformers, dao2023flowmatchinglatentspace}, video generation \citep{davtyan2023efficient, polyak2025moviegencastmedia} and speech \citep{le2023voicebox}. 

A flow is a deterministic, time-continuous, bijective transformation of a $d$-dimensional Euclidean space $
\mathbb{R}^d$, with extensions to other state spaces \citep{campbell2022continuous,gat2024discrete, chen2024flow}. A flow $\psi$ can be defined in terms of a velocity field, often referred to as a vector field, using the following ordinary differential equation (ODE) \citep{lipman2024flowmatchingguidecode}:

\begin{equation}
    \label{eqn:flow_ode}
    \frac{d}{dt} \psi_t = u_t(\psi_t(x)), \ \psi_0(x) = x
\end{equation}

Flow matching is based on learning the underlying velocity field $u_t$. The flow is defined by a process called simulation, i.e. solving the ODE defined in Equation~\ref{eqn:flow_ode} \citep{lipman2024flowmatchingguidecode}. A probability path $p_t$ is required to interpolate between the source, $p_0$, and target distributions, $q$. A common choice for the probability path is the conditional optimal-transport (OT) or linear path with a unit Gaussian source distribution $p_0 = \mathcal{N} (0, I)$.

$$
p_{t|1}(x|x_1) = \mathcal{N}(x|t x_1, (1 - t)^2 I)
$$
    
A neural network is \edit{trained} to approximate the vector field that defines the flow. For Gaussian optimal-transport, the loss function for the velocity field model, $u_t^{\theta}$, parameterized by the model weights $\theta$, is

\begin{equation}
    \mathcal{L}_{\text{CFM}} = \mathbb{E}_{t, X_0, X_1} ||u_t^{\theta}(X_t) - (X_1 - X_0)||^2,
\end{equation}

where $t \sim \mathcal{U}[0,1]$, $X_0 \sim \mathcal{N}(0,1)$ and $X_1 \sim q$. With a trained velocity model, it is then possible to sample from the data distribution by solving the ODE using numerical methods \citep{Iserles_2008}. As the \edit{flow} is bijective, it is also possible to recover the `seed' from the base distribution that generates the target by solving the ODE in reverse from the target distribution.

\textbf{Classifier-free guidance (CFG)}:
Introduced by \cite{ho2022classifierfreediffusionguidance} for diffusion models, classifier-free guidance provides an efficient way of simultaneously approximating conditional and unconditional distributions \citep{zheng2023guided}. This is achieved by replacing guiding information ($y$) used to produce the target with a null vector $\varnothing$ with some probability, $p_{cfg}$. This is a hyperparameter and it is recommended to be set in the range $0.1 \le p_{cfg} \le 0.2$ \citep{ho2022classifierfreediffusionguidance} in order to adequately approximate the unconditional distribution, without affecting the efficacy of the conditional distribution approximation. At inference time, it is then possible to use a weighted combination of the guided and unguided velocities when simulating the ODE \citep{dieleman2022guidance, 2025arXiv250602070H}. 
\begin{equation}
    u_t^{\text{CFG}} = (1 - \omega) \ u_t(x_1|x_t) + \omega \ u_t(x_1|x_t, y)
\end{equation}

where $\omega$ is a hyperparameter that controls the guidance weight.

\section{Approach}
\label{sec: approach}
\begin{figure}
    \centering
    \begin{subfigure}{0.49\linewidth}
        \includegraphics[width=\linewidth]{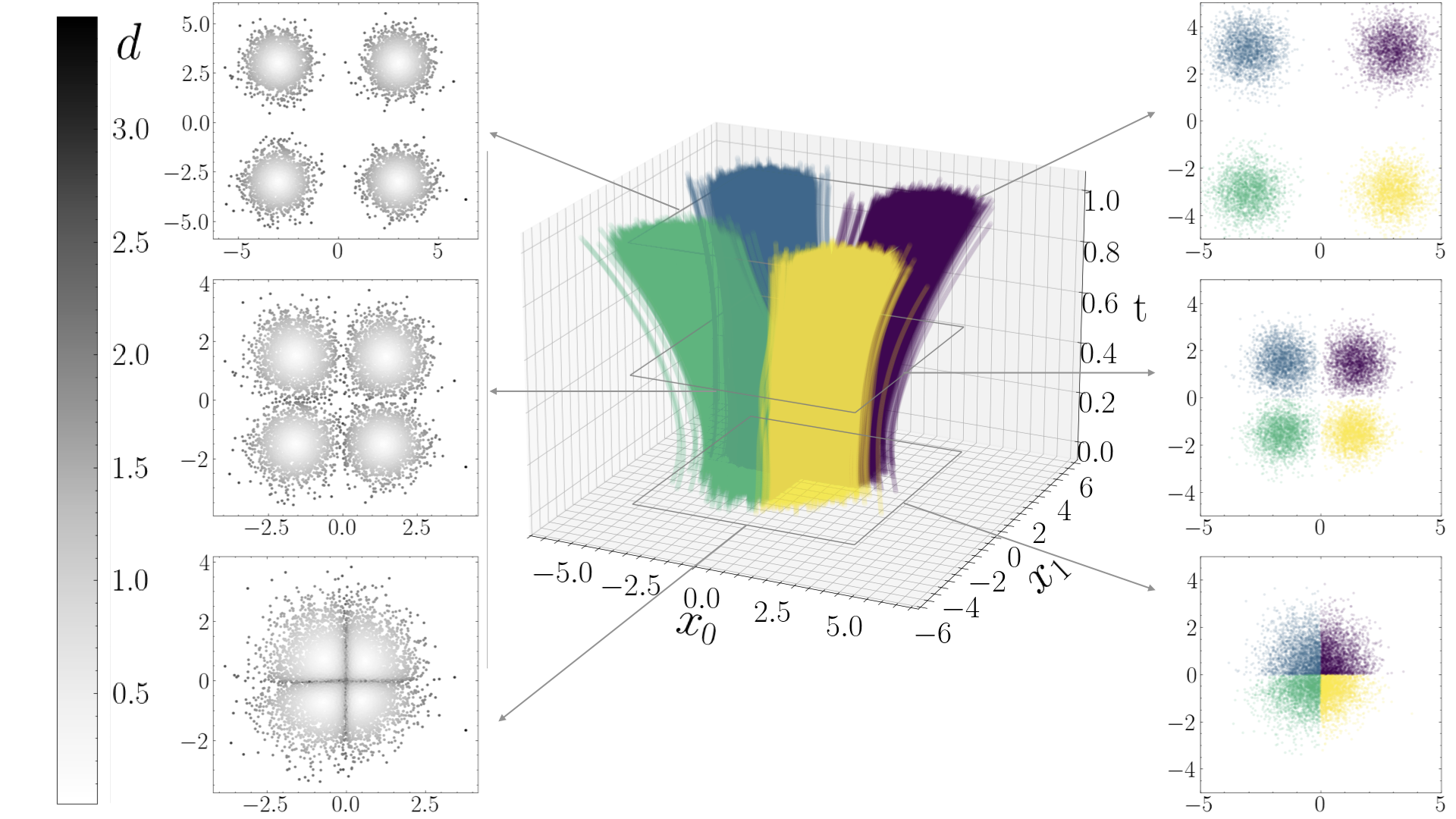}
        \subcaption[]{Unguided flow: Class retrieval is trivial, \\but distance retrieval is not at $t=0$.}
        \label{fig:2d_gaussian_uncond}
    \end{subfigure}
    \begin{subfigure}{0.49\linewidth}
        \includegraphics[width=\linewidth]{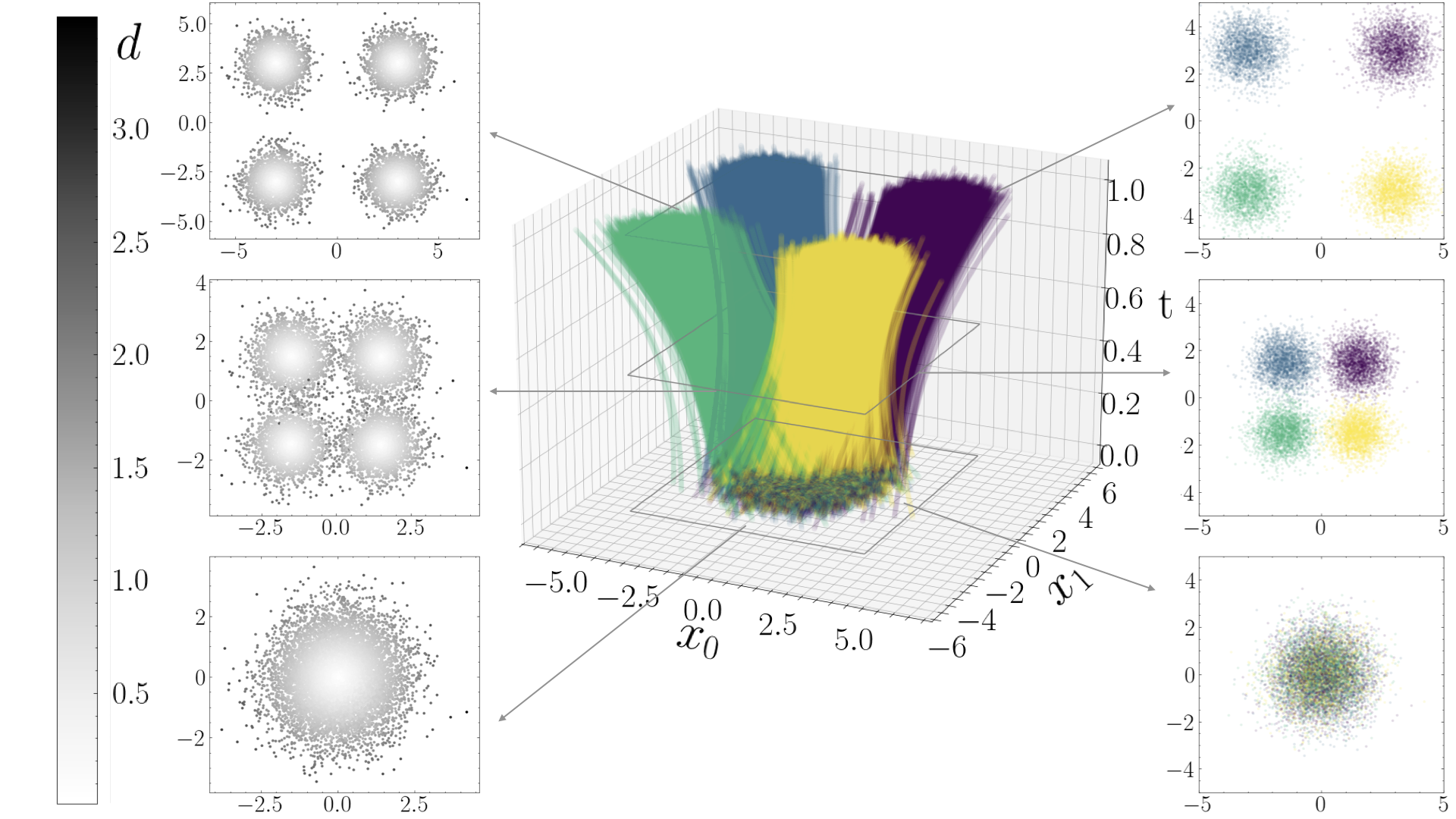}
        \subcaption[]{Class Conditional flow: Distance retrieval is trivial, but class retrieval is not (at $t=0$).}
        \label{fig:2d_gaussian_cond}
    \end{subfigure}
    \caption{
        \edit{Flows of the 2D Gaussian experiment detailed in Section~\ref{subsec: 2d gaussians}. Each figure contains three main features: the central panel shows the flow across time steps for the 2D Gaussian class input data at $t=1$. On either side, three 2D slices at different time steps of the flow are shown to highlight internal structures. The data points in the left slices are colored by their distance to their Gaussian's center as measured at $t=1$. The data in the center and in the right panels are colored by their respective class index belonging to each of the four Gaussians.}
    }
    \label{fig:2d_gaussian_main}
\end{figure}

% Approach outline
WWDC \edit{is an application of flow matching with guidance on existing manifolds of representations. We do not use the latent flow to sample latents in a traditional generative setting. Instead, we embed a sample and flow from it in reverse to the base distribution. Since the flow matching model approximates optimal transport (OT) trajectories \citep{lipman2024flowmatchingguidecode}, the resulting base distribution contains structure from the original manifold, mapped to a Gaussian distribution. For example, Figure~\ref{fig:2d_gaussian_uncond} shows that the class structure present in the target distribution at $t=1$ is preserved in the base distribution at $t=0$ presented in the middle and right panels.}

\edit{We train our flow model with a finite neural network and CFG. This constrained network is optimized to map all conditional and unconditional paths from the base distribution to the target distribution. When reversed, the flow matches features that are consistent across the data distribution to similar portions of the base distribution. Guidance will alter this dynamic for the features used within the conditional guidance. For example, in Figure~\ref{fig:2d_gaussian_cond} the class structure seen in the target distribution at $t=1$ is entirely inaccessible in the base distribution at $t=0$. In cases such as this, where a guiding feature is fully separable, a perfect guided flow would remove the conditional feature from the base distribution. 
}

\edit{Furthermore, we posit that more of the initial manifold's structure is maintained if that original structure matches the base distribution chosen for the flows. In an extreme example, a perfectly unit Gaussian latent distribution would already fulfill the fitting objective of a flow matching OT model with a unit Gaussian base distribution. For this work, we largely investigate VAEs, one of the most widely used neural representation learning models. As in Equation~\ref{eqn:VAE}, the latent space of VAEs are constrained by a KL-loss which (normally) imposes some Gaussianity on the space. We chose our flow's base distributions to be Gaussian. We highlight that, due to the OT condition, the conditional flow minimally distorts structures originally mapped in the VAE's latent manifold.}

\edit{In summary, our selected base and target distributions share properties because the resulting Gaussian base distribution maintains the global structure of the original data manifold.} 
Additionally, due to multiple factors including model limitations and inherent errors in the ODE solution, the flow matching model is \edit{imperfect}. The resulting distribution is therefore not a perfect \edit{unit} Gaussian and some additional local structure may remain.

\textbf{Training procedure}: See Appendix~\ref{app:experiment details} for details on each experiment and training run. For our flow matching approach, we use the Gaussian conditional optimal transport probability path to interpolate between the base and target distributions. With probability $p_{cfg}$, we drop the conditioning information $y$ used in the calculation of the vector field and replace it with a null embedding, $\varnothing$ \citep{ho2022classifierfreediffusionguidance}. This allows us to train and subsequently sample using both guided and unguided flows at inference time.

\textbf{Reverse-flow representations}: The reverse flow is produced by running the ODE solver backwards from the VAE sample at $t=1$ until $t=0$. We adopt the midpoint method \citep{suli_introduction_2003} for simulation at inference in this work. This produces a continuum of representations from the original VAE sample to the base distribution across time steps in the ODE solution.

\section{Results}
\label{subsec: results}
% Motivation for experiment
To verify our expectations of WWDC, we explore three different datasets in increasing order of complexity and decreasing order of experimental control.

\subsection{2D Gaussians}
\label{subsec: 2d gaussians}
% Experiment outline
We seed four synthetic isotropic Gaussians consisting of samples $X\in\mathbb{R}^2$ for our first experiment. For this, we train a simple flow matching model to generate the four Gaussians as the target distribution. The velocity vector field is modeled using a simple multi-layer perceptron (MLP) which takes the position $X$, a time step, and the class information as input.
We use CFG with a null vector of $\varnothing = -1$.

% What it shows
\edit{The resulting flows are presented in} Figure~\ref{fig:2d_gaussian_main}. Figure~\ref{fig:2d_gaussian_uncond} shows the unguided flow, which results in a clear delineation of classes \edit{in the base distribution at $t=0$ (see the right panels)}. We see that when guided on the class label, the class conditioning shows no discernible structure in the base distribution at $t=0$ of Figure~\ref{fig:2d_gaussian_cond}.

\edit{We assign to each point a second feature: the Euclidean distance to the center of its respective Gaussian at $t=1$. In the guided case, Figure~\ref{fig:2d_gaussian_cond}, we see that there exists a very simple map to the distance metric $d$ which is naturally apparent in the base distribution. In contrast, the structure of the classes in the unconditional case, Figure~\ref{fig:2d_gaussian_uncond}, presents a much more complex distance structure.}

\begin{figure}[h!] % TODO: Remove white spaces from the sides of these images
    \centering
    \begin{subfigure}{0.48\linewidth}
        \includegraphics[width=\linewidth]{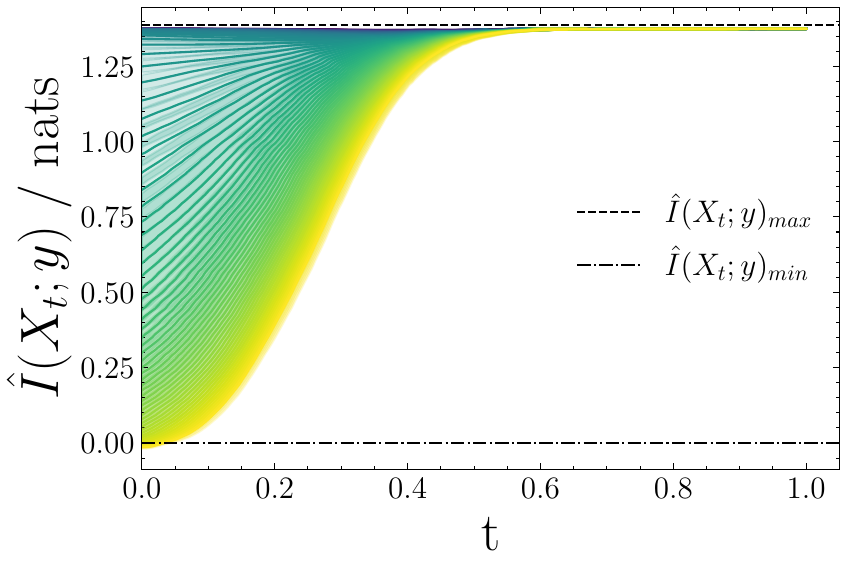}
        \subcaption[]{Evolution of the mutual information between a distribution and the classes.}
        \label{fig:mut_info_guidance}
    \end{subfigure}
    \hfill
    \begin{subfigure}{0.48\linewidth}
        \includegraphics[width=\linewidth]{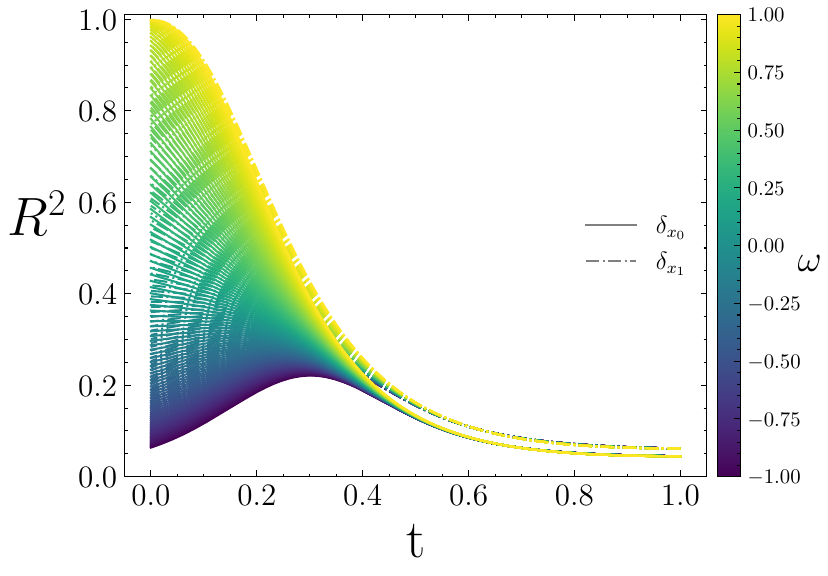}
        \subcaption[]{Linear regressions' $R^2$ scores to dimension-wise distances to the center of the original in $x_0$ and $x_1$.}
        \label{fig:r2_dist}
    \end{subfigure}
    \caption{2D Gaussian evaluations across guidance factors $\omega$ and time $t$. \edit{We see that at weight 1, the secondary feature is surfaced and is fully recoverable in the base distribution and conversely the class mutual information is fully gone.}}
    \label{fig:2}
\end{figure}

\edit{To quantitatively consider these features, we evaluate flows at a variety of guidance weights. The results are presented in Figure~\ref{fig:2}. For the classification problem, we calculate the mutual information with respect to the class of the samples at a given time point. The results are presented in Figure~\ref{fig:mut_info_guidance}, which show that with a guidance weight of $\omega = 1$ (i.e. fully conditional, as presented in Figure~\ref{fig:2d_gaussian_main}), there is a turnover at $t=0.5$ where there is almost full mutual information beyond that time, and progressively less before that time. At the base distribution, $t=0$, there is no mutual information for a guidance weight of 1. Overall, we see that the effect of weighting the guidance is as expected: the weaker the guidance, the more of the class-wise mutual information is preserved.}

\edit{We conduct a similar probe into distance metrics, simplified to a single dimension. We denote the distance from a point to its class's center along one axis or the other as $\delta_{x_i}$. We fit a linear model to predict these features across the times and guidance weights. The results are presented in Figure~\ref{fig:r2_dist}. Because we are using a linear model, the $R^2$ score is effectively zero at $t=1$: a given point's distance cannot be predicted with a linear model, as it is an inherently non-linear setup. However, after guiding to $t=0$ with a weight of 1, equivalent to Figure~\ref{fig:2d_gaussian_cond}, we see increasingly better variance explainability by the simple linear model over these features until, at $t=0$, the full dimension-wise distance can be mapped linearly. Here the unguided, $\omega=0$, case results in an $R^2\approx0.3$. This is expected as the linear explainability has certainly increased in comparison to a non-linear problem, but many samples are not correctly mapped as the resulting space is warped by the importance of class-conditional information, which has not been removed.}

\subsection{cMNIST}
\label{subsec: cmnist}
As a more complex experiment, we consider colored MNIST \citep{6296535} (cMNIST). \edit{We generate} a random RGB value $(r, g, b) \in [0.05,0.95]^3$ and multiply a three-channel version of the MNIST data by this value. We withhold $b$ from the guidance so that we can use it as a secondary feature of interest and still examine the properties of $r$ and $g$.

% Experiment - CMNIST + Why
We first train a $\beta$-VAE on cMNIST with minimal weighting on the KL penalty weight, \edit{$\beta = 1 \times 10^{-4}$}, and a latent of size 64 for improved generation quality. We then train a flow matching model on the latent space of the VAE using a simple MLP to parameterize the velocity vector field. This model is conditioned on the digit class and the maximum red and green values of the colored digit, while blue is withheld from the conditioning. Label dropout is used to estimate the unguided velocity field.

\begin{figure}[h]
    \centering
    % Row 1
    \begin{subfigure}[b]{0.48\linewidth}
        \centering
        \includegraphics[height=0.825\linewidth]{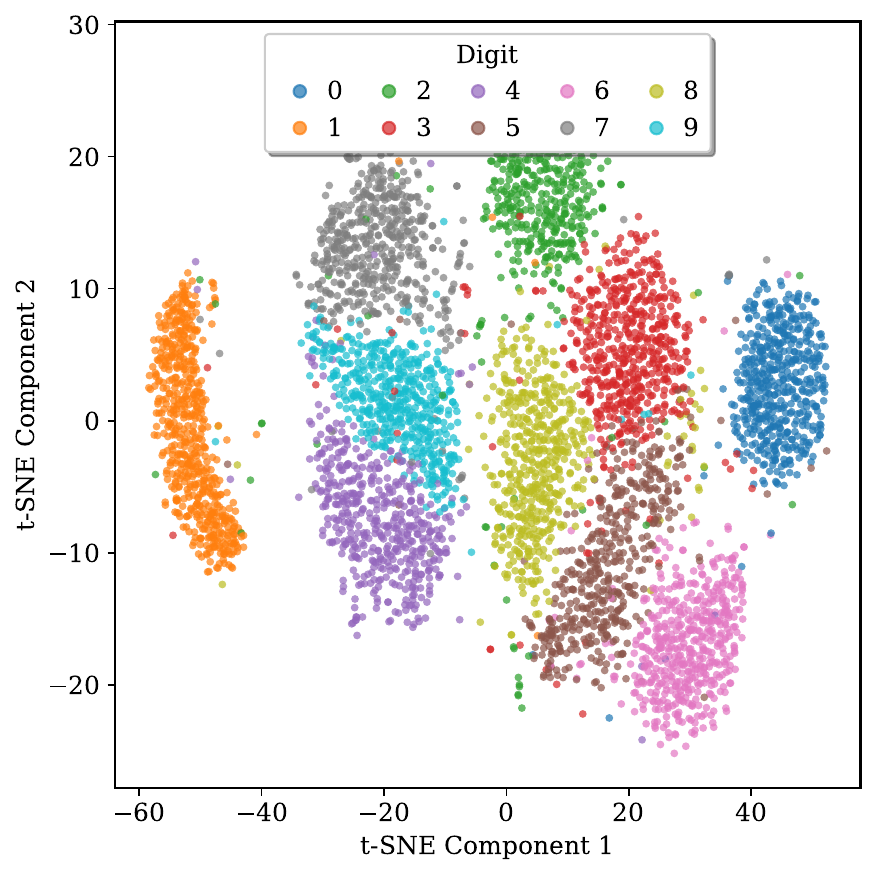}
        \caption{VAE ($t=1$) - Class}
        \label{fig:cmnist_tsne_vae}
    \end{subfigure}
    \hfill
    \begin{subfigure}[b]{0.48\linewidth}
        \centering
        \includegraphics[width=\linewidth]{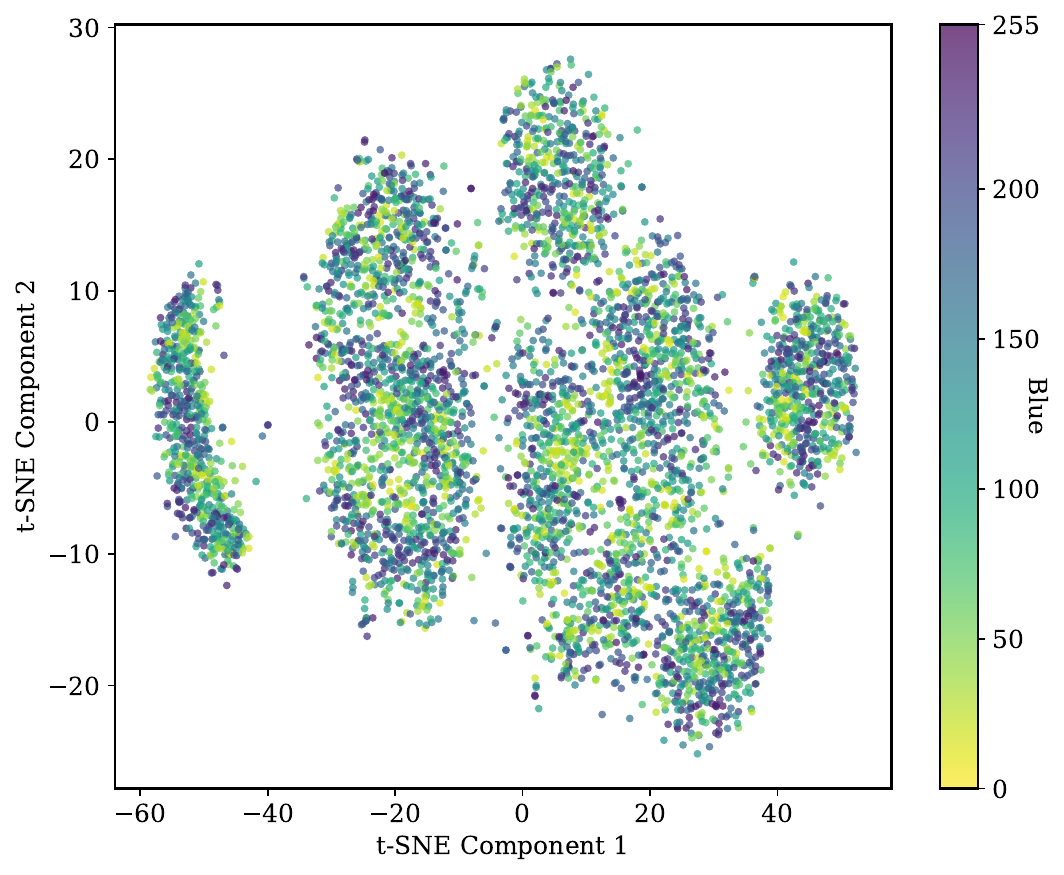}
        \caption{\edit{VAE ($t=1$) - Blue}}
        \label{fig:cmnist_tsne_vae_blue}
    \end{subfigure}

    \vspace{0.5em}

    % Row 2
    \begin{subfigure}[b]{0.48\linewidth}
        \centering
        \includegraphics[height=0.825\linewidth]{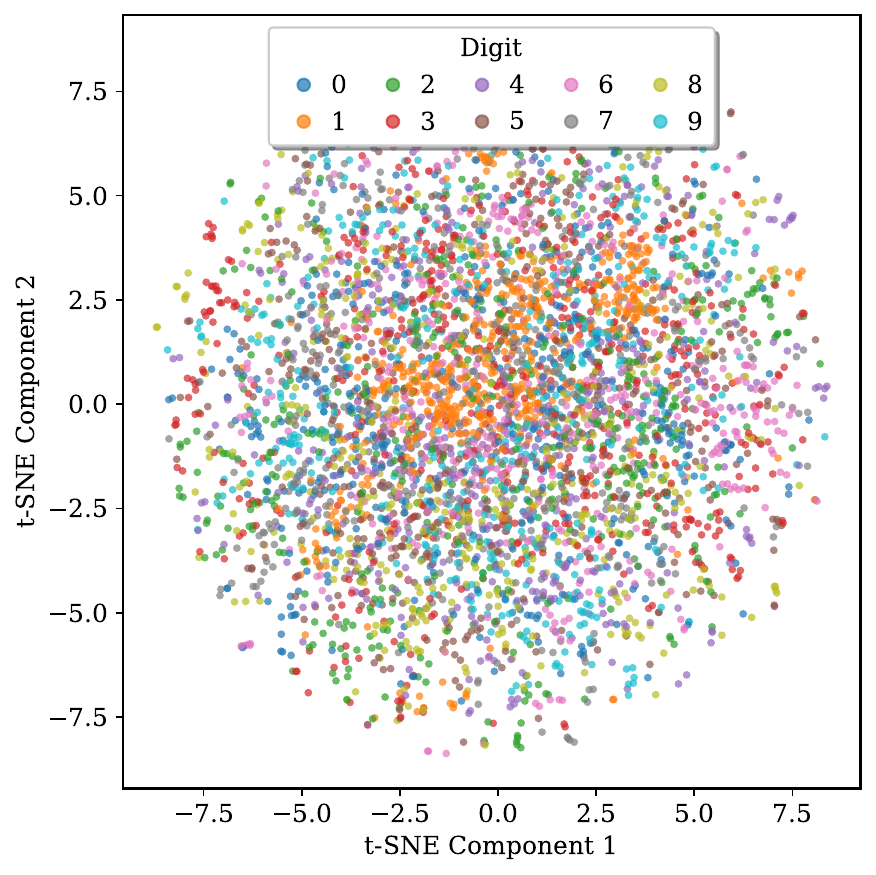}
        \caption{Guided ($t=0$) - Class}
        \label{fig:cmnist_tsne_cond}
    \end{subfigure}
    \hfill
    \begin{subfigure}[b]{0.48\linewidth}
        \centering
        \includegraphics[width=\linewidth]{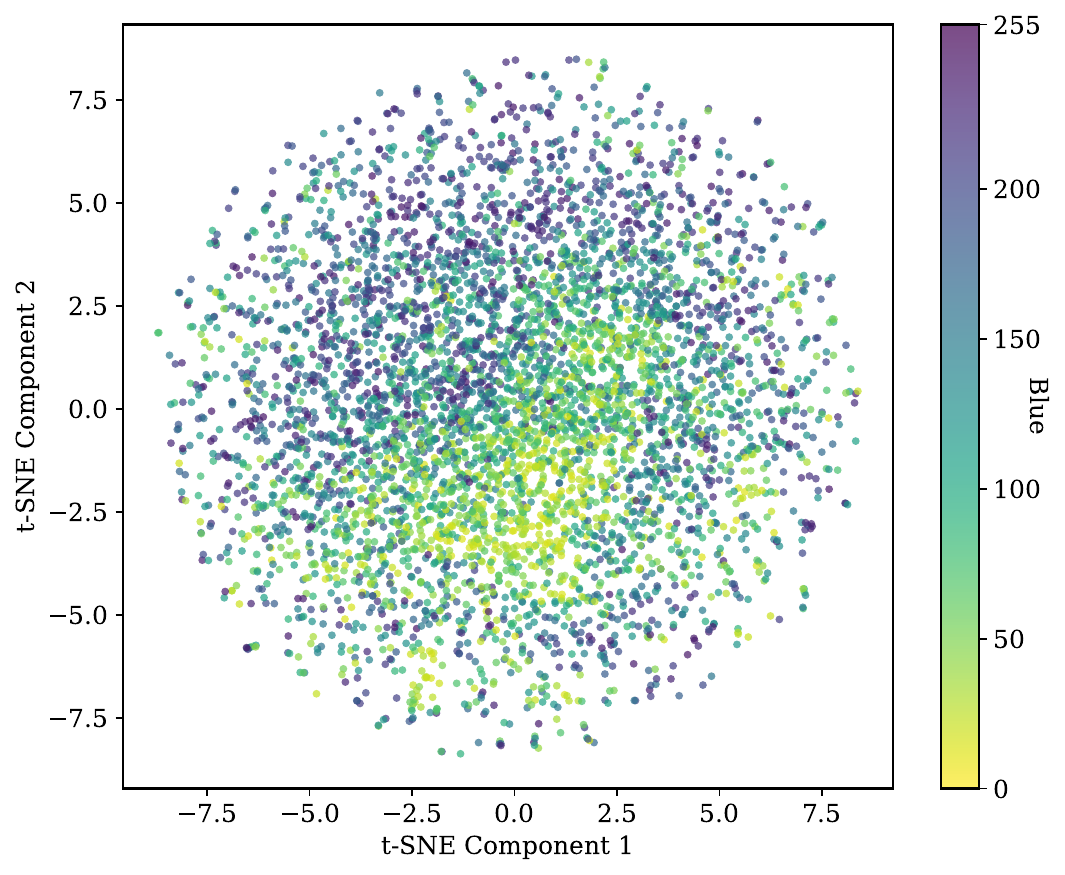}
        \caption{Guided ($t=0$) - Blue}
        \label{fig:cmnist_tsne_cond_blue}
    \end{subfigure}

    \caption{t-SNE projections of cMNIST embeddings. The VAE space in
    \subref{fig:cmnist_tsne_vae} is dominated by class structure \edit{making the task of identifying subtle features, such as blue in \subref{fig:cmnist_tsne_vae_blue}, difficult.} The reverse-guided
    flow in \subref{fig:cmnist_tsne_cond} removes the most visible class structure while
    preserving the underlying color, as shown in \subref{fig:cmnist_tsne_cond_blue}, which
    surfaces more clearly in the structure of the new space \edit{than in \subref{fig:cmnist_tsne_vae_blue}.}}
    \label{fig:cmnist_tsne}
\end{figure}

% Results on embedding space, classification and regression
We use t-SNE \citep{JMLR:v9:vandermaaten08a} (see Appendix~\ref{app:cmnist} for more details and figures) to visualize 2D projections of the latent spaces. These are presented in Figure~\ref{fig:cmnist_tsne_vae}, where we show the class of each point, and the resulting projection shows clear structure. We guide based on red, green, and digit class. The resulting $t=0$ space is projected with the same t-SNE hyperparameters and is shown in Figure~\ref{fig:cmnist_tsne_cond}. We see the class structure almost entirely disappear. We note we don't expect it to disappear entirely, as there are confounding features which are informative over class, for example: consider how expressive `straightness' could be for predicting ones.

In tandem, we plot the same structures but colored by the strength of the blue color in the samples, a feature we didn't condition on. Here we see some, but clearly no obvious structure in the original VAE space in Figure~\ref{fig:cmnist_tsne_vae_blue}. In comparison, the guided case, presented in Figure~\ref{fig:cmnist_tsne_cond_blue}, presents a gradient across the space. It is unlikely that an annotator would find structure in the VAE space, but it is plausible that an annotator could have found the blue feature within the base conditional space. 
% fig: tsne class + non class 

\begin{figure}
    \begin{subfigure}{0.486\linewidth}
        \includegraphics[width=\linewidth]{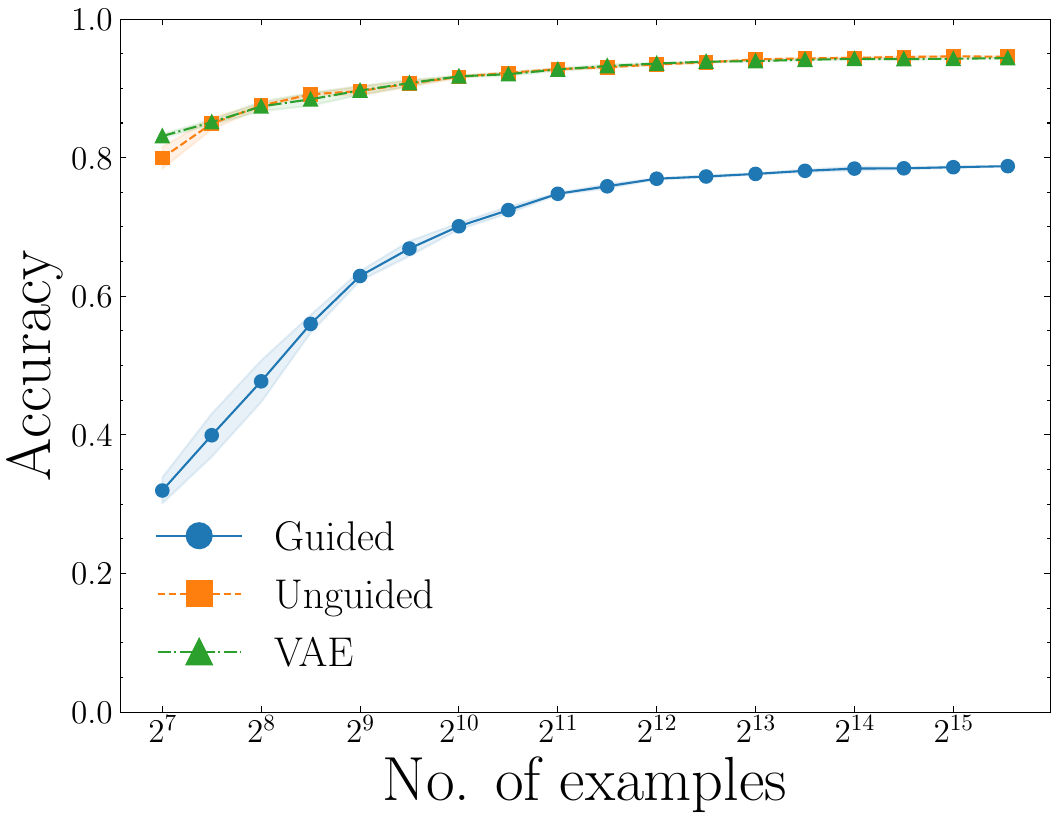}
        \caption{Linear probe classification accuracy.}
        \label{fig:mnist linear probe classification}
    \end{subfigure}
    \begin{subfigure}{0.483\linewidth}
        \includegraphics[width=\linewidth]{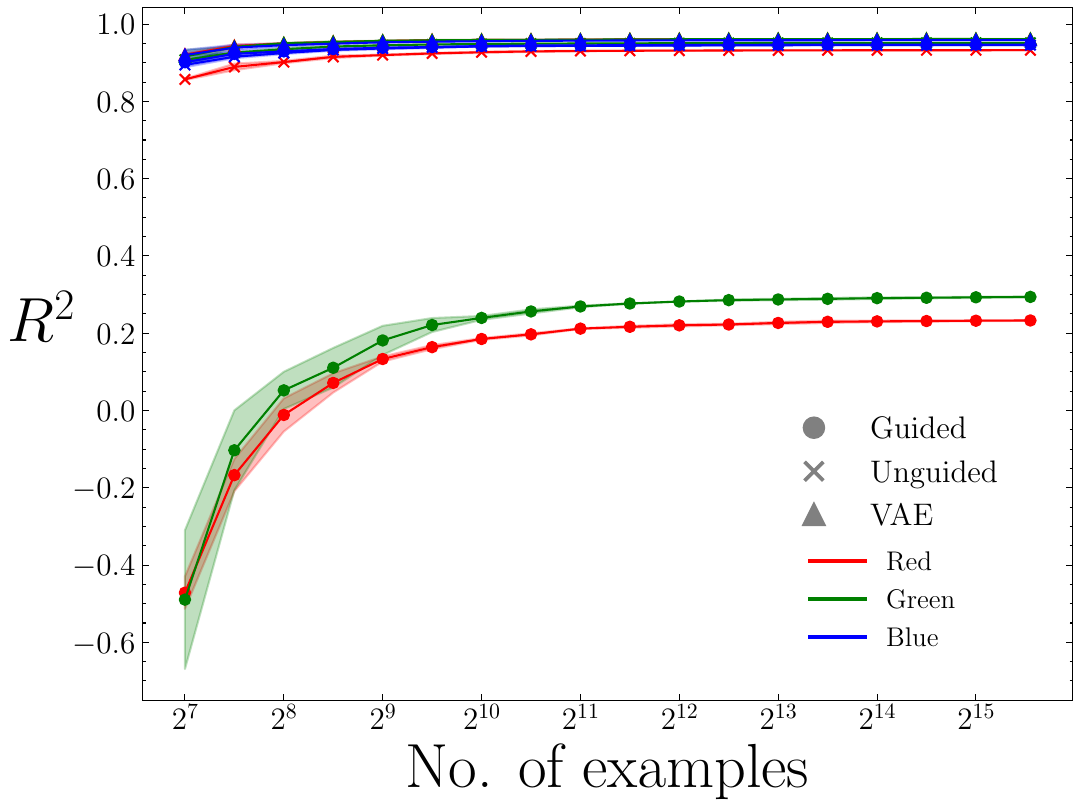}
        \caption{Linear probe color regression $R^2$-score.}
        \label{fig:mnist linear probe regression}
    \end{subfigure}
    \caption[]{Linear probe evaluations for classification and regression tasks on the colored MNIST task. Note that the blue value is withheld and is consistently recovered throughout both flows.}
    \label{fig:mnist linear probes}
\end{figure}

To evaluate how guidance suppresses the digit, red, and green features, and how this enables recovery of blue, we train simple linear regression models \citep{scikit-learn} on the representations. We vary the number of training examples used in these models and show the results in Figure~\ref{fig:mnist linear probes}. Figure~\ref{fig:mnist linear probe classification} shows that the digit classification accuracy drops significantly with guidance in comparison to unguided and VAE representations. Similarly, in the case of a linear model regressing to the $r$ and $g$ values, the ability to recover the guiding information is repressed, and especially with fewer training samples. We see that the guidance signal does not adversely affect the sample size required to recover blue, i.e. the feature outside of the guidance, highlighting how even in this space, a few labeled samples can be quite useful.

We note that the accuracy and $R^2$ scores not dropping by a larger degree than they have, is likely due to two factors. Firstly, as previously noted, the digit class information is inherently bound to other features that are not unique to that digit, and therefore cannot be removed with a simple class annotation (e.g. straightness or number of loops). Secondly, as we wanted high quality samples, our latent space is not as restricted as is possible for MNIST reconstruction. This was an intentional decision to enable feature interpretability and is also likely why having more training samples improves recovery of guided properties given the scale of the latent space. \edit{See training details in Appendix~\ref{app:cmnist}.}

\begin{figure}[h!]
\centering
\includegraphics[width=\linewidth]{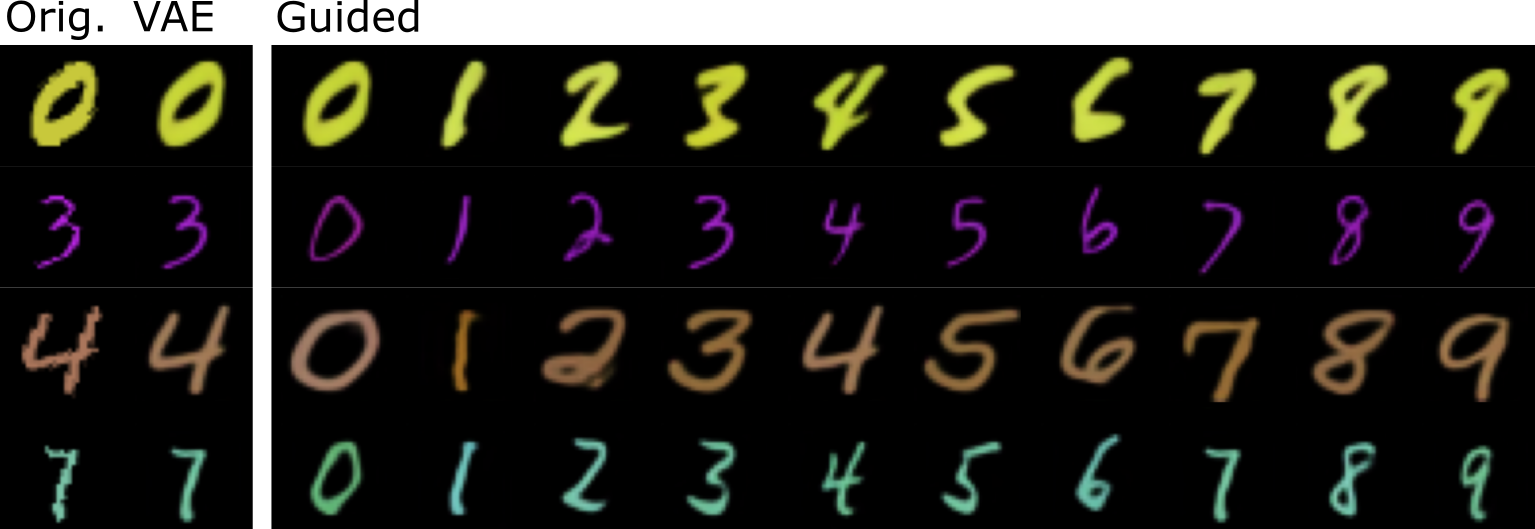}
\caption{Style transfer in colored MNIST: The guided $t=0$ embeddings are used to initialize a flow model but the guidance is switched, conditioning on another digit, to produce stylistically consistent digits in the VAE space. See Figure~\ref{fig:guided_cfg_transfer} in the appendix for more examples.}
\label{fig:cmnist_style_transfer_small}
\end{figure}

% Style transfer process
Our interest in high quality reconstructions is because, as a tool, WWDC also natively enables sample generation. We first move backwards in the flow using the digit conditioning to $t=0$, and then initialize the flow forwards using different guidance. Figure~\ref{fig:cmnist_style_transfer_small} shows results of this for cMNIST samples and class changes. This enables relatively cheap inspection of synthetic data. We envision this as a supporting function for the WWDC annotation loop, outlined in Fig~\ref{fig:wwdc}, as one can conceivably inspect the same sample across different guidance to spot semantic similarities. For example, in the cMNIST case of Figure~\ref{fig:cmnist_style_transfer_small}  we note stroke widths (e.g. 0 vs. 3), digit position (e.g. 4 vs. 7), and, of course, color are all preserved across guided samples.
%
% Plot of style transfer to highlight wwdc.

\subsection{Galaxy10}
\label{subsec: galaxy10}
\begin{figure}[h!]
    \centering
    \includegraphics[width=\linewidth]{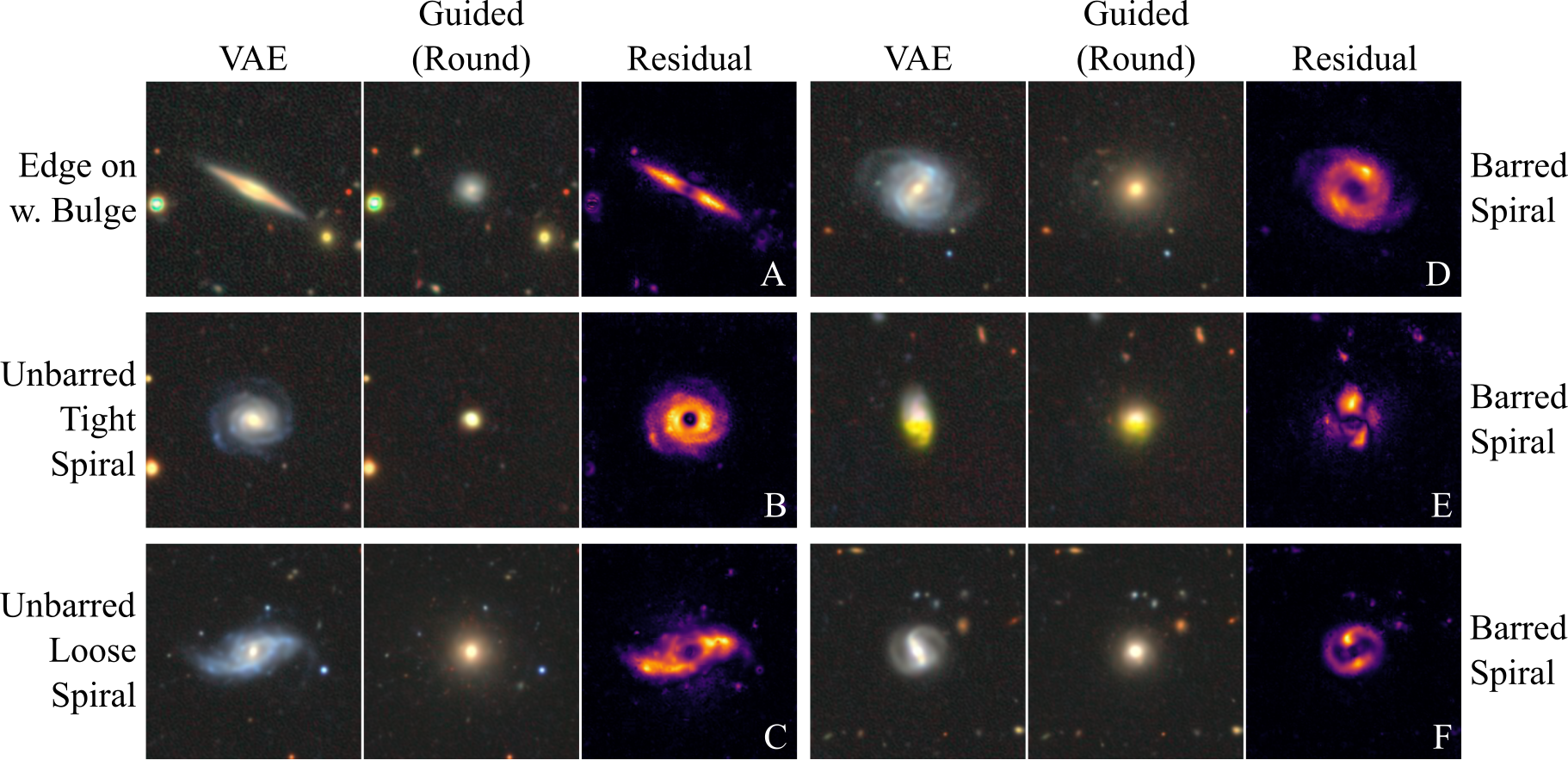}
    \caption{Samples of visual feature isolation for Galaxy10. Galaxies, their guidance-generated `round' versions, the residuals between these images are shown. Features associated with the original galaxies are clearly separated from the remaining features of the image. For more examples, see Figure~\ref{fig:pair_barred_vs_tight} and Figure~\ref{fig:pair_loose_vs_additional} in the appendix.}
    \label{fig:galaxy10_feature_iso}
\end{figure}
We now explore how techniques from previous sections are applicable in a more realistic setting. We consider scientific data, and specifically the Galaxy 10 DECaLS dataset \citep{2019MNRAS.483.3255L}. Galaxies contained in the data are separated into ten broad but not necessarily distinct morphological classes. These include: disturbed, merging, round smooth, in-between round smooth, cigar round smooth, barred spiral, unbarred tight spiral, unbarred loose spiral, edge-on without bulge and edge-on with bulge. We train a VAE with a minimal \edit{KL-weighting of $\beta = 1 \times 10^{-6}$} and with latents $z \in \mathbb{R}^{4 \times 32 \times 32}$ \edit{to enable high quality reconstructions}.

To demonstrate how class and concepts are disentangled during the latent flow, we first select different classes from the data and project to the base distribution using $\omega=1$. We then flow from the base towards $t=1$ using the `round' class as guiding information with $\omega=3.5$ to increase the guidance signal, following \citet{ho2022classifierfreediffusionguidance}. We choose `round' as it is the least semantically complex structure and so enables us to view residuals of what the process has changed about the image without introducing other features from a different galaxy class. Results are presented in Figure~\ref{fig:galaxy10_feature_iso}. We note that the background features remain unchanged in the forward-guided image, confirming that the model has identified the galaxy of interest from extremely simple class labels and the guidance leaves these features unchanged. In sample E, we also note that the lower half of the galaxy is yellow, which is not a physical feature but rather an imaging artifact. This feature is also preserved in the forward generated galaxy despite the structure of the galaxy being changed.
Through the residuals, we see the isolation of the guided features in that image. This is a native property of WWDC and \edit{could be used} to understand what exactly has been captured (or not captured) by a presumptive class or measurement, \edit{adding to the scientific discovery loop envisioned in Figure~\ref{fig:wwdc}.} Importantly, separating galaxy features like this will directly enable analyses in astrophysics through modern \edit{surveys } \citep[e.g. LSST;][]{2019ApJ...873..111I} \edit{where high-dimensional complex features are abundant and whose exploration is limited largely by cost.}

\section{Conclusion}
\label{sec: Conclusion}
This work introduced What We Don't C (WWDC) as an approach to manifold disentanglement for structured discovery. Disentanglement being a distinctly difficult task, we relax the constraints on the problem to disentangling a known signal from a structured manifold with the purpose of enabling access to other features within the manifold.

We demonstrated with a toy Gaussian problem how flow models can produce \textit{meaningful representations} of the data that \edit{surface} secondary features of interest while suppressing known information.
The colored MNIST experiment shows this in a more complex setting where we selectively retrieve color from latents \textit{only} if we do not condition on them. We highlight that the latents must contain information on the style of a given digit by guided generation. Finally, we apply the method to astrophysical data, where we show that our model can disentangle important class features in galaxy images. 

Critically, this approach enables reuse of existing VAEs and we expect adoption of this approach to enable structured discovery and annotation through representation learning. We identify a clear immediate use case for WWDC in assisting researchers \edit{to} explore what information they haven't yet captured, either because they didn't think of it, or could not access easily.

\section*{Acknowledgement of LLM Usage}
% The authors thank their funding bodies. 
Large language models, including Gemini and ChatGPT, were used during the research of this paper for source finding and literature exploration. Additionally, AI assisted development tools including Cursor were used in the development of the code used in this manuscript.

\bibliography{iclr2026_conference}

@inproceedings{Liang2018CollaborativeFiltering,
author = {Liang, Dawen and Krishnan, Rahul G. and Hoffman, Matthew D. and Jebara, Tony},
title = {Variational Autoencoders for Collaborative Filtering},
year = {2018},
isbn = {9781450356398},
publisher = {International World Wide Web Conferences Steering Committee},
address = {Republic and Canton of Geneva, CHE},
url = {https://doi.org/10.1145/3178876.3186150},
doi = {10.1145/3178876.3186150},
booktitle = {Proceedings of the 2018 World Wide Web Conference},
pages = {689–698},
numpages = {10},
location = {Lyon, France},
series = {WWW '18}
}

@inproceedings{
luo2023diffusion,
title={Diffusion Hyperfeatures: Searching Through Time and Space for Semantic Correspondence},
author={Grace Luo and Lisa Dunlap and Dong Huk Park and Aleksander Holynski and Trevor Darrell},
booktitle={Thirty-seventh Conference on Neural Information Processing Systems},
year={2023},
url={https://openreview.net/forum?id=Vm1zeYqwdc}
}

@inproceedings{
    tang2023emergent,
    title={Emergent Correspondence from Image Diffusion},
    author={Luming Tang and Menglin Jia and Qianqian Wang and Cheng Perng Phoo and Bharath Hariharan},
    booktitle={Thirty-seventh Conference on Neural Information Processing Systems},
    year={2023},
    url={https://openreview.net/forum?id=ypOiXjdfnU}
}

@article{oquab2023dinov2,
  title={Dinov2: Learning robust visual features without supervision},
  author={Oquab, Maxime and Darcet, Timoth{\'e}e and Moutakanni, Th{\'e}o and Vo, Huy and Szafraniec, Marc and Khalidov, Vasil and Fernandez, Pierre and Haziza, Daniel and Massa, Francisco and El-Nouby, Alaaeldin and others},
  journal={arXiv preprint arXiv:2304.07193},
  year={2023}
}

@article{He21MAE,
  author       = {Kaiming He and
                  Xinlei Chen and
                  Saining Xie and
                  Yanghao Li and
                  Piotr Doll{\'{a}}r and
                  Ross B. Girshick},
  title        = {Masked Autoencoders Are Scalable Vision Learners},
  journal      = {CoRR},
  volume       = {abs/2111.06377},
  year         = {2021},
  url          = {https://arxiv.org/abs/2111.06377},
  eprinttype    = {arXiv},
  eprint       = {2111.06377},
}

@article{tao2024llms-afc, 
  year     = {2024}, 
  title    = {{LLMs} are Also Effective Embedding Models: An In-depth Overview}, 
  author   = {Tao, Chongyang and Shen, Tao and Gao, Shen and Zhang, Junshuo and Li, Zhen and Tao, Zhengwei and Ma, Shuai}, 
  journal  = {{arXiv}}, 
  doi      = {10.48550/arxiv.2412.12591}, 
  eprint   = {2412.12591}, 
}

@InProceedings{JimenezRezende2014StochasticBA,
  title = 	 {Stochastic Backpropagation and Approximate Inference in Deep Generative Models},
  author = 	 {Rezende, Danilo Jimenez and Mohamed, Shakir and Wierstra, Daan},
  booktitle = 	 {Proceedings of the 31st International Conference on Machine Learning},
  pages = 	 {1278--1286},
  year = 	 {2014},
  editor = 	 {Xing, Eric P. and Jebara, Tony},
  volume = 	 {32},
  series = 	 {Proceedings of Machine Learning Research},
  address = 	 {Bejing, China},
  month = 	 {22--24 Jun},
  publisher =    {PMLR},
  url = 	 {https://proceedings.mlr.press/v32/rezende14.html}
}

@inproceedings{
higgins2017betavae,
title={beta-{VAE}: Learning Basic Visual Concepts with a Constrained Variational Framework},
author={Irina Higgins and Loic Matthey and Arka Pal and Christopher Burgess and Xavier Glorot and Matthew Botvinick and Shakir Mohamed and Alexander Lerchner},
booktitle={International Conference on Learning Representations},
year={2017},
url={https://openreview.net/forum?id=Sy2fzU9gl}
}

@misc{burgess2018understandingdisentanglingbetavae,
      title={Understanding disentangling in $\beta$-VAE}, 
      author={Christopher P. Burgess and Irina Higgins and Arka Pal and Loic Matthey and Nick Watters and Guillaume Desjardins and Alexander Lerchner},
      year={2018},
      eprint={1804.03599},
      archivePrefix={arXiv},
      primaryClass={stat.ML},
      url={https://arxiv.org/abs/1804.03599}, 
}

@inproceedings{chen2019isolatingsourcesdisentanglementvariational,
 author = {Chen, Ricky T. Q. and Li, Xuechen and Grosse, Roger B and Duvenaud, David K},
 booktitle = {Advances in Neural Information Processing Systems},
 editor = {S. Bengio and H. Wallach and H. Larochelle and K. Grauman and N. Cesa-Bianchi and R. Garnett},
 pages = {},
 publisher = {Curran Associates, Inc.},
 title = {Isolating Sources of Disentanglement in Variational Autoencoders},
 url = {https://proceedings.neurips.cc/paper_files/paper/2018/file/1ee3dfcd8a0645a25a35977997223d22-Paper.pdf},
 volume = {31},
 year = {2018}
}

@misc{kingma2022autoencodingvariationalbayes,
      title={Auto-Encoding Variational Bayes}, 
      author={Diederik P Kingma and Max Welling},
      year={2022},
      eprint={1312.6114},
      archivePrefix={arXiv},
      primaryClass={stat.ML},
      url={https://arxiv.org/abs/1312.6114}, 
}

@inproceedings{
ho2022classifierfreediffusionguidance,
title={Classifier-Free Diffusion Guidance},
author={Jonathan Ho and Tim Salimans},
booktitle={NeurIPS 2021 Workshop on Deep Generative Models and Downstream Applications},
year={2021},
url={https://openreview.net/forum?id=qw8AKxfYbI}
}

@inproceedings{siddharth2017learningdisentangledrepresentationssemisupervised,
  title={Learning Disentangled Representations with Semi-Supervised Deep Generative Models},
  author={Siddharth, N and Paige, Brooks and van de Meent, Jan-Willem and Desmaison, Alban and Goodman, Noah and Kohli, Pushmeet and Wood, Frank and Torr, Philip HS},
  booktitle={Thirty-first Annual Conference on Neural Information Processing Systems},
  pages={5925--5935},
  year={2017},
  organization={Neural Information Processing Systems}
}

@misc{lipman2024flowmatchingguidecode,
      title={Flow Matching Guide and Code}, 
      author={Yaron Lipman and Marton Havasi and Peter Holderrieth and Neta Shaul and Matt Le and Brian Karrer and Ricky T. Q. Chen and David Lopez-Paz and Heli Ben-Hamu and Itai Gat},
      year={2024},
      eprint={2412.06264},
      archivePrefix={arXiv},
      primaryClass={cs.LG},
      url={https://arxiv.org/abs/2412.06264}, 
}

@misc{cheung2015discoveringhiddenfactorsvariation,
      title={Discovering Hidden Factors of Variation in Deep Networks}, 
      author={Brian Cheung and Jesse A. Livezey and Arjun K. Bansal and Bruno A. Olshausen},
      year={2015},
      eprint={1412.6583},
      archivePrefix={arXiv},
      primaryClass={cs.LG},
      url={https://arxiv.org/abs/1412.6583}, 
}

@inproceedings{
lipman2023flowmatchinggenerativemodeling,
title={Flow Matching for Generative Modeling},
author={Yaron Lipman and Ricky T. Q. Chen and Heli Ben-Hamu and Maximilian Nickel and Matthew Le},
booktitle={The Eleventh International Conference on Learning Representations },
year={2023},
url={https://openreview.net/forum?id=PqvMRDCJT9t}
}

@ARTICLE{2025arXiv250602070H,
       author = {{Holderrieth}, Peter and {Erives}, Ezra},
        title = "{An Introduction to Flow Matching and Diffusion Models}",
      journal = {arXiv e-prints},
         year = 2025,
        month = jun,
          eid = {arXiv:2506.02070},
        pages = {arXiv:2506.02070},
          doi = {10.48550/arXiv.2506.02070},
archivePrefix = {arXiv},
       eprint = {2506.02070},
 primaryClass = {stat.ML},
       adsurl = {https://ui.adsabs.harvard.edu/abs/2025arXiv250602070H}
}

@inproceedings{NIPS2015_8d55a249,
 author = {Sohn, Kihyuk and Lee, Honglak and Yan, Xinchen},
 booktitle = {Advances in Neural Information Processing Systems},
 editor = {C. Cortes and N. Lawrence and D. Lee and M. Sugiyama and R. Garnett},
 pages = {},
 publisher = {Curran Associates, Inc.},
 title = {Learning Structured Output Representation using Deep Conditional Generative Models},
 url = {https://proceedings.neurips.cc/paper_files/paper/2015/file/8d55a249e6baa5c06772297520da2051-Paper.pdf},
 volume = {28},
 year = {2015}
}

@ARTICLE{6296535,
  author={Deng, Li},
  journal={IEEE Signal Processing Magazine}, 
  title={The MNIST Database of Handwritten Digit Images for Machine Learning Research [Best of the Web]}, 
  year={2012},
  volume={29},
  number={6},
  pages={141-142},
  doi={10.1109/MSP.2012.2211477}}

@inproceedings{perez2017filmvisualreasoninggeneral,
  title={Film: Visual reasoning with a general conditioning layer},
  author={Perez, Ethan and Strub, Florian and De Vries, Harm and Dumoulin, Vincent and Courville, Aaron},
  booktitle={Proceedings of the AAAI conference on artificial intelligence},
  volume={32},
  year={2018}
}

@article{scikit-learn,
  title={Scikit-learn: Machine Learning in {P}ython},
  author={Pedregosa, F. and Varoquaux, G. and Gramfort, A. and Michel, V.
          and Thirion, B. and Grisel, O. and Blondel, M. and Prettenhofer, P.
          and Weiss, R. and Dubourg, V. and Vanderplas, J. and Passos, A. and
          Cournapeau, D. and Brucher, M. and Perrot, M. and Duchesnay, E.},
  journal={Journal of Machine Learning Research},
  volume={12},
  pages={2825--2830},
  year={2011}
}

@ARTICLE{2011MNRAS.410..166L,
       author = {{Lintott}, Chris and {Schawinski}, Kevin and {Bamford}, Steven and {Slosar}, An{\r{a}}{\textthreequarters}e and {Land}, Kate and {Thomas}, Daniel and {Edmondson}, Edd and {Masters}, Karen and {Nichol}, Robert C. and {Raddick}, M. Jordan and {Szalay}, Alex and {Andreescu}, Dan and {Murray}, Phil and {Vandenberg}, Jan},
        title = "{Galaxy Zoo 1: data release of morphological classifications for nearly 900 000 galaxies}",
      journal = {MNRAS},
         year = 2011,
        month = jan,
       volume = {410},
       number = {1},
        pages = {166-178},
          doi = {10.1111/j.1365-2966.2010.17432.x},
archivePrefix = {arXiv},
       eprint = {1007.3265},
 primaryClass = {astro-ph.GA},
       adsurl = {https://ui.adsabs.harvard.edu/abs/2011MNRAS.410..166L}
}

@ARTICLE{2019MNRAS.483.3255L,
       author = {{Leung}, Henry W. and {Bovy}, Jo},
        title = "{Deep learning of multi-element abundances from high-resolution spectroscopic data}",
      journal = {MNRAS},
         year = 2019,
        month = mar,
       volume = {483},
       number = {3},
        pages = {3255-3277},
          doi = {10.1093/mnras/sty3217},
archivePrefix = {arXiv},
       eprint = {1808.04428},
 primaryClass = {astro-ph.GA},
       adsurl = {https://ui.adsabs.harvard.edu/abs/2019MNRAS.483.3255L}
}

@misc{von-platen-etal-2022-diffusers,
  author = {Patrick von Platen and Suraj Patil and Anton Lozhkov and Pedro Cuenca and Nathan Lambert and Kashif Rasul and Mishig Davaadorj and Dhruv Nair and Sayak Paul and William Berman and Yiyi Xu and Steven Liu and Thomas Wolf},
  title = {Diffusers: State-of-the-art diffusion models},
  year = {2022},
  publisher = {GitHub},
  journal = {GitHub repository},
  howpublished = {\url{https://github.com/huggingface/diffusers}}
}

@inproceedings{2015arXiv150504597R,
  title={U-net: Convolutional networks for biomedical image segmentation},
  author={Ronneberger, Olaf and Fischer, Philipp and Brox, Thomas},
  booktitle={International Conference on Medical image computing and computer-assisted intervention},
  pages={234--241},
  year={2015},
  organization={Springer}
}

@article{tong2024improving,
title={Improving and generalizing flow-based generative models with minibatch optimal transport},
author={Alexander Tong and Kilian FATRAS and Nikolay Malkin and Guillaume Huguet and Yanlei Zhang and Jarrid Rector-Brooks and Guy Wolf and Yoshua Bengio},
journal={Transactions on Machine Learning Research},
issn={2835-8856},
year={2024},
url={https://openreview.net/forum?id=CD9Snc73AW},
note={Expert Certification}
}

@inproceedings{tong2023simulation,
  title={Simulation-free Schr{\"o}dinger bridges via score and flow matching},
  author={Tong, Alexander and Malkin, Nikolay and Fatras, Kilian and Atanackovic, Lazar and Zhang, Yanlei and Huguet, Guillaume and Wolf, Guy and Bengio, Yoshua},
  booktitle={The 27th International Conference on Artificial Intelligence and Statistics},
  pages={1279--1287},
  year={2024},
  organization={Journal of Machine Learning Research-Proceedings Track}
}

@ARTICLE{2019ApJ...873..111I,
       author = {{Ivezi{\'c}}, {\v{Z}}eljko and {Kahn}, Steven M. and {Tyson}, J. Anthony and {Abel}, Bob and {Acosta}, Emily and {Allsman}, Robyn and {Alonso}, David and {AlSayyad}, Yusra and {Anderson}, Scott F. and {Andrew}, John and {Angel}, James Roger P. and {Angeli}, George Z. and {Ansari}, Reza and {Antilogus}, Pierre and {Araujo}, Constanza and {Armstrong}, Robert and {Arndt}, Kirk T. and {Astier}, Pierre and {Aubourg}, {\'E}ric and {Auza}, Nicole and {Axelrod}, Tim S. and {Bard}, Deborah J. and {Barr}, Jeff D. and {Barrau}, Aurelian and {Bartlett}, James G. and {Bauer}, Amanda E. and {Bauman}, Brian J. and {Baumont}, Sylvain and {Bechtol}, Ellen and {Bechtol}, Keith and {Becker}, Andrew C. and {Becla}, Jacek and {Beldica}, Cristina and {Bellavia}, Steve and {Bianco}, Federica B. and {Biswas}, Rahul and {Blanc}, Guillaume and {Blazek}, Jonathan and {Blandford}, Roger D. and {Bloom}, Josh S. and {Bogart}, Joanne and {Bond}, Tim W. and {Booth}, Michael T. and {Borgland}, Anders W. and {Borne}, Kirk and {Bosch}, James F. and {Boutigny}, Dominique and {Brackett}, Craig A. and {Bradshaw}, Andrew and {Brandt}, William Nielsen and {Brown}, Michael E. and {Bullock}, James S. and {Burchat}, Patricia and {Burke}, David L. and {Cagnoli}, Gianpietro and {Calabrese}, Daniel and {Callahan}, Shawn and {Callen}, Alice L. and {Carlin}, Jeffrey L. and {Carlson}, Erin L. and {Chandrasekharan}, Srinivasan and {Charles-Emerson}, Glenaver and {Chesley}, Steve and {Cheu}, Elliott C. and {Chiang}, Hsin-Fang and {Chiang}, James and {Chirino}, Carol and {Chow}, Derek and {Ciardi}, David R. and {Claver}, Charles F. and {Cohen-Tanugi}, Johann and {Cockrum}, Joseph J. and {Coles}, Rebecca and {Connolly}, Andrew J. and {Cook}, Kem H. and {Cooray}, Asantha and {Covey}, Kevin R. and {Cribbs}, Chris and {Cui}, Wei and {Cutri}, Roc and {Daly}, Philip N. and {Daniel}, Scott F. and {Daruich}, Felipe and {Daubard}, Guillaume and {Daues}, Greg and {Dawson}, William and {Delgado}, Francisco and {Dellapenna}, Alfred and {de Peyster}, Robert and {de Val-Borro}, Miguel and {Digel}, Seth W. and {Doherty}, Peter and {Dubois}, Richard and {Dubois-Felsmann}, Gregory P. and {Durech}, Josef and {Economou}, Frossie and {Eifler}, Tim and {Eracleous}, Michael and {Emmons}, Benjamin L. and {Fausti Neto}, Angelo and {Ferguson}, Henry and {Figueroa}, Enrique and {Fisher-Levine}, Merlin and {Focke}, Warren and {Foss}, Michael D. and {Frank}, James and {Freemon}, Michael D. and {Gangler}, Emmanuel and {Gawiser}, Eric and {Geary}, John C. and {Gee}, Perry and {Geha}, Marla and {Gessner}, Charles J.~B. and {Gibson}, Robert R. and {Gilmore}, D. Kirk and {Glanzman}, Thomas and {Glick}, William and {Goldina}, Tatiana and {Goldstein}, Daniel A. and {Goodenow}, Iain and {Graham}, Melissa L. and {Gressler}, William J. and {Gris}, Philippe and {Guy}, Leanne P. and {Guyonnet}, Augustin and {Haller}, Gunther and {Harris}, Ron and {Hascall}, Patrick A. and {Haupt}, Justine and {Hernandez}, Fabio and {Herrmann}, Sven and {Hileman}, Edward and {Hoblitt}, Joshua and {Hodgson}, John A. and {Hogan}, Craig and {Howard}, James D. and {Huang}, Dajun and {Huffer}, Michael E. and {Ingraham}, Patrick and {Innes}, Walter R. and {Jacoby}, Suzanne H. and {Jain}, Bhuvnesh and {Jammes}, Fabrice and {Jee}, M. James and {Jenness}, Tim and {Jernigan}, Garrett and {Jevremovi{\'c}}, Darko and {Johns}, Kenneth and {Johnson}, Anthony S. and {Johnson}, Margaret W.~G. and {Jones}, R. Lynne and {Juramy-Gilles}, Claire and {Juri{\'c}}, Mario and {Kalirai}, Jason S. and {Kallivayalil}, Nitya J. and {Kalmbach}, Bryce and {Kantor}, Jeffrey P. and {Karst}, Pierre and {Kasliwal}, Mansi M. and {Kelly}, Heather and {Kessler}, Richard and {Kinnison}, Veronica and {Kirkby}, David and {Knox}, Lloyd and {Kotov}, Ivan V. and {Krabbendam}, Victor L. and {Krughoff}, K. Simon and {Kub{\'a}nek}, Petr and {Kuczewski}, John and {Kulkarni}, Shri and {Ku}, John and {Kurita}, Nadine R. and {Lage}, Craig S. and {Lambert}, Ron and {Lange}, Travis and {Langton}, J. Brian and {Le Guillou}, Laurent and {Levine}, Deborah and {Liang}, Ming and {Lim}, Kian-Tat and {Lintott}, Chris J. and {Long}, Kevin E. and {Lopez}, Margaux and {Lotz}, Paul J. and {Lupton}, Robert H. and {Lust}, Nate B. and {MacArthur}, Lauren A. and {Mahabal}, Ashish and {Mandelbaum}, Rachel and {Markiewicz}, Thomas W. and {Marsh}, Darren S. and {Marshall}, Philip J. and {Marshall}, Stuart and {May}, Morgan and {McKercher}, Robert and {McQueen}, Michelle and {Meyers}, Joshua and {Migliore}, Myriam and {Miller}, Michelle and {Mills}, David J.},
        title = "{LSST: From Science Drivers to Reference Design and Anticipated Data Products}",
      journal = {ApJ},
         year = 2019,
        month = mar,
       volume = {873},
       number = {2},
          eid = {111},
        pages = {111},
          doi = {10.3847/1538-4357/ab042c},
archivePrefix = {arXiv},
       eprint = {0805.2366},
 primaryClass = {astro-ph},
       adsurl = {https://ui.adsabs.harvard.edu/abs/2019ApJ...873..111I}
}

@misc{dao2023flowmatchinglatentspace,
      title={Flow Matching in Latent Space}, 
      author={Quan Dao and Hao Phung and Binh Nguyen and Anh Tran},
      year={2023},
      eprint={2307.08698},
      archivePrefix={arXiv},
      primaryClass={cs.CV},
      url={https://arxiv.org/abs/2307.08698}, 
}

@misc{polyak2025moviegencastmedia,
      title={Movie Gen: A Cast of Media Foundation Models}, 
      author={Adam Polyak and Amit Zohar and Andrew Brown and Andros Tjandra and Animesh Sinha and Ann Lee and Apoorv Vyas and Bowen Shi and Chih-Yao Ma and Ching-Yao Chuang and David Yan and Dhruv Choudhary and Dingkang Wang and Geet Sethi and Guan Pang and Haoyu Ma and Ishan Misra and Ji Hou and Jialiang Wang and Kiran Jagadeesh and Kunpeng Li and Luxin Zhang and Mannat Singh and Mary Williamson and Matt Le and Matthew Yu and Mitesh Kumar Singh and Peizhao Zhang and Peter Vajda and Quentin Duval and Rohit Girdhar and Roshan Sumbaly and Sai Saketh Rambhatla and Sam Tsai and Samaneh Azadi and Samyak Datta and Sanyuan Chen and Sean Bell and Sharadh Ramaswamy and Shelly Sheynin and Siddharth Bhattacharya and Simran Motwani and Tao Xu and Tianhe Li and Tingbo Hou and Wei-Ning Hsu and Xi Yin and Xiaoliang Dai and Yaniv Taigman and Yaqiao Luo and Yen-Cheng Liu and Yi-Chiao Wu and Yue Zhao and Yuval Kirstain and Zecheng He and Zijian He and Albert Pumarola and Ali Thabet and Artsiom Sanakoyeu and Arun Mallya and Baishan Guo and Boris Araya and Breena Kerr and Carleigh Wood and Ce Liu and Cen Peng and Dimitry Vengertsev and Edgar Schonfeld and Elliot Blanchard and Felix Juefei-Xu and Fraylie Nord and Jeff Liang and John Hoffman and Jonas Kohler and Kaolin Fire and Karthik Sivakumar and Lawrence Chen and Licheng Yu and Luya Gao and Markos Georgopoulos and Rashel Moritz and Sara K. Sampson and Shikai Li and Simone Parmeggiani and Steve Fine and Tara Fowler and Vladan Petrovic and Yuming Du},
      year={2025},
      eprint={2410.13720},
      archivePrefix={arXiv},
      primaryClass={cs.CV},
      url={https://arxiv.org/abs/2410.13720}, 
}

@inproceedings{esser2024scalingrectifiedflowtransformers,
  author={Patrick Esser and Sumith Kulal and Andreas Blattmann and Rahim Entezari and Jonas Müller and Harry Saini and Yam Levi and Dominik Lorenz and Axel Sauer and Frederic Boesel and Dustin Podell and Tim Dockhorn and Zion English and Robin Rombach},
  title={Scaling Rectified Flow Transformers for High-Resolution Image Synthesis},
  year={2024},
  cdate={1704067200000},
  url={https://openreview.net/forum?id=FPnUhsQJ5B},
  booktitle={ICML},
}

@article{zheng2023guided,
  publtype={informal},
  author={Qinqing Zheng and Matt Le and Neta Shaul and Yaron Lipman and Aditya Grover and Ricky T. Q. Chen},
  title={Guided Flows for Generative Modeling and Decision Making},
  year={2023},
  cdate={1672531200000},
  journal={CoRR},
  volume={abs/2311.13443},
  url={https://doi.org/10.48550/arXiv.2311.13443}
}

@inproceedings{10.5555/3666122.3668288, author = {Song, Yue and Keller, T. Anderson and Sebe, Nicu and Welling, Max}, title = {Flow factorized representation learning}, year = {2023}, publisher = {Curran Associates Inc.}, address = {Red Hook, NY, USA}, booktitle = {Proceedings of the 37th International Conference on Neural Information Processing Systems}, articleno = {2166}, numpages = {22}, location = {New Orleans, LA, USA}, series = {NIPS '23} }

@article{fuest2024diffusion,
  title={Diffusion models and representation learning: A survey},
  author={Fuest, Michael and Ma, Pingchuan and Gui, Ming and Schusterbauer, Johannes and Hu, Vincent Tao and Ommer, Bjorn},
  journal={arXiv preprint arXiv:2407.00783},
  year={2024}
}

@inproceedings{davtyan2023efficient,
  title={Efficient video prediction via sparsely conditioned flow matching},
  author={Davtyan, Aram and Sameni, Sepehr and Favaro, Paolo},
  booktitle={Proceedings of the IEEE/CVF International Conference on Computer Vision},
  pages={23263--23274},
  year={2023}
}

@article{le2023voicebox,
  title={Voicebox: Text-guided multilingual universal speech generation at scale},
  author={Le, Matthew and Vyas, Apoorv and Shi, Bowen and Karrer, Brian and Sari, Leda and Moritz, Rashel and Williamson, Mary and Manohar, Vimal and Adi, Yossi and Mahadeokar, Jay and others},
  journal={Advances in neural information processing systems},
  volume={36},
  pages={14005--14034},
  year={2023}
}

@article{bengio2013representation,
  title={Representation learning: A review and new perspectives},
  author={Bengio, Yoshua and Courville, Aaron and Vincent, Pascal},
  journal={IEEE transactions on pattern analysis and machine intelligence},
  volume={35},
  number={8},
  pages={1798--1828},
  year={2013},
  publisher={IEEE}
}

@inproceedings{DBLP:conf/iclr/0001SB18,
  author       = {Abhishek Kumar and
                  Prasanna Sattigeri and
                  Avinash Balakrishnan},
  title        = {Variational Inference of Disentangled Latent Concepts from Unlabeled
                  Observations},
  booktitle    = {6th International Conference on Learning Representations, {ICLR} 2018,
                  Vancouver, BC, Canada, April 30 - May 3, 2018, Conference Track Proceedings},
  publisher    = {OpenReview.net},
  year         = {2018},
  url          = {https://openreview.net/forum?id=H1kG7GZAW},
  bibsource    = {dblp computer science bibliography, https://dblp.org}
}

@inproceedings{jeong2019learning,
  title={Learning discrete and continuous factors of data via alternating disentanglement},
  author={Jeong, Yeonwoo and Song, Hyun Oh},
  booktitle={International Conference on Machine Learning},
  pages={3091--3099},
  year={2019},
  organization={PMLR}
}

@inproceedings{shao2020controlvae,
  title={Controlvae: Controllable variational autoencoder},
  author={Shao, Huajie and Yao, Shuochao and Sun, Dachun and Zhang, Aston and Liu, Shengzhong and Liu, Dongxin and Wang, Jun and Abdelzaher, Tarek},
  booktitle={International conference on machine learning},
  pages={8655--8664},
  year={2020},
  organization={PMLR}
}

@inproceedings{locatello2019challenging,
  title={Challenging common assumptions in the unsupervised learning of disentangled representations},
  author={Locatello, Francesco and Bauer, Stefan and Lucic, Mario and Raetsch, Gunnar and Gelly, Sylvain and Sch{\"o}lkopf, Bernhard and Bachem, Olivier},
  booktitle={international conference on machine learning},
  pages={4114--4124},
  year={2019},
  organization={PMLR}
}

@article{kingma2014semi,
  title={Semi-supervised learning with deep generative models},
  author={Kingma, Diederik P and Rezende, Danilo J and Mohamed, Shakir and Welling, Max},
  journal={Advances in neural information processing systems},
  volume={27},
  year={2014}
}

@inproceedings{NIPS2015_ced556cd,
 author = {Kulkarni, Tejas D and Whitney, William F. and Kohli, Pushmeet and Tenenbaum, Josh},
 booktitle = {Advances in Neural Information Processing Systems},
 editor = {C. Cortes and N. Lawrence and D. Lee and M. Sugiyama and R. Garnett},
 pages = {},
 publisher = {Curran Associates, Inc.},
 title = {Deep Convolutional Inverse Graphics Network},
 url = {https://proceedings.neurips.cc/paper_files/paper/2015/file/ced556cd9f9c0c8315cfbe0744a3baf0-Paper.pdf},
 volume = {28},
 year = {2015}
}

@article{
doi:10.1126/science.1127647,
author = {G. E. Hinton  and R. R. Salakhutdinov },
title = {Reducing the Dimensionality of Data with Neural Networks},
journal = {Science},
volume = {313},
number = {5786},
pages = {504-507},
year = {2006},
doi = {10.1126/science.1127647},
URL = {https://www.science.org/doi/abs/10.1126/science.1127647},
eprint = {https://www.science.org/doi/pdf/10.1126/science.1127647}}

@inproceedings{
chen2024flow,
title={Flow Matching on General Geometries},
author={Ricky T. Q. Chen and Yaron Lipman},
booktitle={The Twelfth International Conference on Learning Representations},
year={2024},
url={https://openreview.net/forum?id=g7ohDlTITL}
}

@article{Pearson01111901,
author = {Karl Pearson},
title = {LIII. On lines and planes of closest fit to systems of points in space },
journal = {The London, Edinburgh, and Dublin Philosophical Magazine and Journal of Science},
volume = {2},
number = {11},
pages = {559--572},
year = {1901},
publisher = {Taylor \& Francis},
doi = {10.1080/14786440109462720},


URL = { 
    
        https://doi.org/10.1080/14786440109462720
    
    

},
eprint = { 
    
        https://doi.org/10.1080/14786440109462720
}
}

@article{JMLR:v9:vandermaaten08a,
  author  = {Laurens van der Maaten and Geoffrey Hinton},
  title   = {Visualizing Data using t-SNE},
  journal = {Journal of Machine Learning Research},
  year    = {2008},
  volume  = {9},
  number  = {86},
  pages   = {2579--2605},
  url     = {http://jmlr.org/papers/v9/vandermaaten08a.html}
}

@inproceedings{chen2020simple,
  title={A simple framework for contrastive learning of visual representations},
  author={Chen, Ting and Kornblith, Simon and Norouzi, Mohammad and Hinton, Geoffrey},
  booktitle={International conference on machine learning},
  pages={1597--1607},
  year={2020},
  organization={PmLR}
}

@article{grill2020bootstrap,
  title={Bootstrap your own latent-a new approach to self-supervised learning},
  author={Grill, Jean-Bastien and Strub, Florian and Altch{\'e}, Florent and Tallec, Corentin and Richemond, Pierre and Buchatskaya, Elena and Doersch, Carl and Avila Pires, Bernardo and Guo, Zhaohan and Gheshlaghi Azar, Mohammad and others},
  journal={Advances in neural information processing systems},
  volume={33},
  pages={21271--21284},
  year={2020}
}

@inproceedings{assran2023self,
  title={Self-supervised learning from images with a joint-embedding predictive architecture},
  author={Assran, Mahmoud and Duval, Quentin and Misra, Ishan and Bojanowski, Piotr and Vincent, Pascal and Rabbat, Michael and LeCun, Yann and Ballas, Nicolas},
  booktitle={Proceedings of the IEEE/CVF Conference on Computer Vision and Pattern Recognition},
  pages={15619--15629},
  year={2023}
}

@article{campbell2022continuous,
  title={A continuous time framework for discrete denoising models},
  author={Campbell, Andrew and Benton, Joe and De Bortoli, Valentin and Rainforth, Thomas and Deligiannidis, George and Doucet, Arnaud},
  journal={Advances in Neural Information Processing Systems},
  volume={35},
  pages={28266--28279},
  year={2022}
}

@article{gat2024discrete,
  title={Discrete flow matching},
  author={Gat, Itai and Remez, Tal and Shaul, Neta and Kreuk, Felix and Chen, Ricky TQ and Synnaeve, Gabriel and Adi, Yossi and Lipman, Yaron},
  journal={Advances in Neural Information Processing Systems},
  volume={37},
  pages={133345--133385},
  year={2024}
}

@book{Iserles_2008, place={Cambridge}, edition={2}, series={Cambridge Texts in Applied Mathematics}, title={A First Course in the Numerical Analysis of Differential Equations}, publisher={Cambridge University Press}, author={Iserles, Arieh}, year={2008}, collection={Cambridge Texts in Applied Mathematics}}

@misc{dieleman2022guidance,
  author = {Dieleman, Sander},
  title = {Guidance: a cheat code for diffusion models},
  url = {https://benanne.github.io/2022/05/26/guidance.html},
  year = {2022}
}

@article{kullback1951information,
  title={On information and sufficiency},
  author={Kullback, Solomon and Leibler, Richard A},
  journal={The annals of mathematical statistics},
  volume={22},
  number={1},
  pages={79--86},
  year={1951},
  publisher={JSTOR}
}

@inproceedings{khattab2020colbert,
  title={Colbert: Efficient and effective passage search via contextualized late interaction over bert},
  author={Khattab, Omar and Zaharia, Matei},
  booktitle={Proceedings of the 43rd International ACM SIGIR conference on research and development in Information Retrieval},
  pages={39--48},
  year={2020}
}

@article{ren2024deep,
  title={Deep clustering: A comprehensive survey},
  author={Ren, Yazhou and Pu, Jingyu and Yang, Zhimeng and Xu, Jie and Li, Guofeng and Pu, Xiaorong and Yu, Philip S and He, Lifang},
  journal={IEEE transactions on neural networks and learning systems},
  volume={36},
  number={4},
  pages={5858--5878},
  year={2024},
  publisher={IEEE}
}

@inproceedings{
YM.2020Self-labelling,
title={Self-labelling via simultaneous clustering and representation learning},
author={Asano, YM and Rupprecht, C and Vedaldi, A},
booktitle={International Conference on Learning Representations},
year={2020},
url={https://openreview.net/forum?id=Hyx-jyBFPr}
}

@article{han2022adbench,
  title={Adbench: Anomaly detection benchmark},
  author={Han, Songqiao and Hu, Xiyang and Huang, Hailiang and Jiang, Minqi and Zhao, Yue},
  journal={Advances in neural information processing systems},
  volume={35},
  pages={32142--32159},
  year={2022}
}

@article{mcinnes2018umap,
  title={Umap: Uniform manifold approximation and projection for dimension reduction},
  author={McInnes, Leland and Healy, John and Melville, James},
  journal={arXiv preprint arXiv:1802.03426},
  year={2018}
}

@book{suli_introduction_2003,
	series = {An {Introduction} to {Numerical} {Analysis}},
	title = {An {Introduction} to {Numerical} {Analysis}},
	isbn = {978-0-521-00794-8},
	url = {https://books.google.co.uk/books?id=hj9weaqJTbQC},
	publisher = {Cambridge University Press},
	author = {Süli, E. and Mayers, D.F.},
	year = {2003},
	lccn = {2003273860},
}

@article{balestriero2025lejepa,
  title={LeJEPA: Provable and Scalable Self-Supervised Learning Without the Heuristics},
  author={Balestriero, Randall and LeCun, Yann},
  journal={arXiv preprint arXiv:2511.08544},
  year={2025}
}

@misc{yue2026imagegenerationsphereencoder,
      title={Image Generation with a Sphere Encoder}, 
      author={Kaiyu Yue and Menglin Jia and Ji Hou and Tom Goldstein},
      year={2026},
      eprint={2602.15030},
      archivePrefix={arXiv},
      primaryClass={cs.CV},
      url={https://arxiv.org/abs/2602.15030}, 
}

@misc{kumar2026learningmanifoldunlockingstandard,
      title={Learning on the Manifold: Unlocking Standard Diffusion Transformers with Representation Encoders}, 
      author={Amandeep Kumar and Vishal M. Patel},
      year={2026},
      eprint={2602.10099},
      archivePrefix={arXiv},
      primaryClass={cs.LG},
      url={https://arxiv.org/abs/2602.10099}, 
}

@misc{mvparakhin_ml_tidbits_wristband,
  author       = {Parakhin, Mikhail},
  title        = {Wristband Gaussian Loss — ml-tidbits},
  year         = {2026},
  howpublished = {\url{https://github.com/mvparakhin/ml-tidbits/blob/main/docs/wristband.md}},
  organization = {GitHub},
  url          = {https://github.com/mvparakhin/ml-tidbits/blob/main/docs/wristband.md}
}

@inproceedings{
betser2026infonce,
title={Info{NCE} Induces Gaussian Distribution},
author={Roy Betser and Eyal Gofer and Meir Yossef Levi and Guy Gilboa},
booktitle={The Fourteenth International Conference on Learning Representations},
year={2026},
url={https://openreview.net/forum?id=BlSH7gNQSq}
}

@misc{ye2025distributionmatchingvariationalautoencoder,
      title={Distribution Matching Variational AutoEncoder}, 
      author={Sen Ye and Jianning Pei and Mengde Xu and Shuyang Gu and Chunyu Wang and Liwei Wang and Han Hu},
      year={2025},
      eprint={2512.07778},
      archivePrefix={arXiv},
      primaryClass={cs.CV},
      url={https://arxiv.org/abs/2512.07778}, 
}
\bibliographystyle{iclr2026_conference}

\appendix

\section{Experiment Details}
\label{app:experiment details}
\subsection{2d Gaussians}
\label{app:2d_gauss}

The four Gaussians were each generated with covariances $\Sigma = 0.5 I$ and respective means from the set $\{(\pm3,\pm3)^\intercal, (\pm3,\mp3)^\intercal\}\in\mathbb{R}$.
The model that is fit to the flow is an MLP consisting of four linear layers with ELU activations in-between, resulting in 8.5k trainable parameters. We used a label dropout probability of $p_{cfg}=0.2$.

\subsection{cMNIST}
\label{app:cmnist}
We augment the MNIST dataset \cite{6296535} with color information to provide a highly controllable experiment in order to verify the intuitions we have built from the 2D Gaussian case in section \ref{subsec: 2d gaussians}. The color information is determined such that 

\begin{equation}
    c_i \sim \mathcal{U}(0.05, 0.95)
\end{equation}

for $i=1,2,3$ (i.e. the R, G and B channels of the image) and $\mathcal{U}$ denotes the uniform distribution. We also use the following data augmentations:

\begin{itemize}
    \item Rotations: $\theta \sim \mathcal{N}(0, 10)$
    \item Scale: $s \sim \mathcal{N}(1, 0.1)$
\end{itemize}

The $\beta$-VAE consists of CNN encoder and decoder. The encoder and decoders are inspired by the VGG architecture (convolutions followed by batch normalization layers, ReLU, and max pooling to step down in scale). The decoder uses up-sampling instead of max pooling. There are also linear projections with ReLU activations from the CNN encoder to the latent space and a linear projection from the latent space to the CNN decoder. This model contains 23.4 M parameters.

The model is trained using $\beta = 1 \times 10^{-4}$. The flow model consists of a simple MLP of four hidden layers with $GeLU$ activations taking the latent vector and time as inputs. This model consists of 171 k parameters. The intermediate linear layers are modulated with a projection of the class information. For this, the digit is embedded and concatenated with the maximum red and green value. We then project the concatenated information using a linear layer to output a scale and shift terms for the intermediate states of the network after linear layers in a method similar to \cite{perez2017filmvisualreasoninggeneral}. Guidance dropout is used with a probability of $p_{cfg}=0.1$.

The t-SNE plots were generated with the following parameters \cite{scikit-learn}: we used two components, perplexity was set to 100 and maximum iterations was 5000. The random state was 42 with the 'auto' learning rate setting and it was initialized using the 'pca' option.

\begin{figure}[h]
    \centering
    \begin{subfigure}{0.52\linewidth}
        \includegraphics[width=\linewidth]{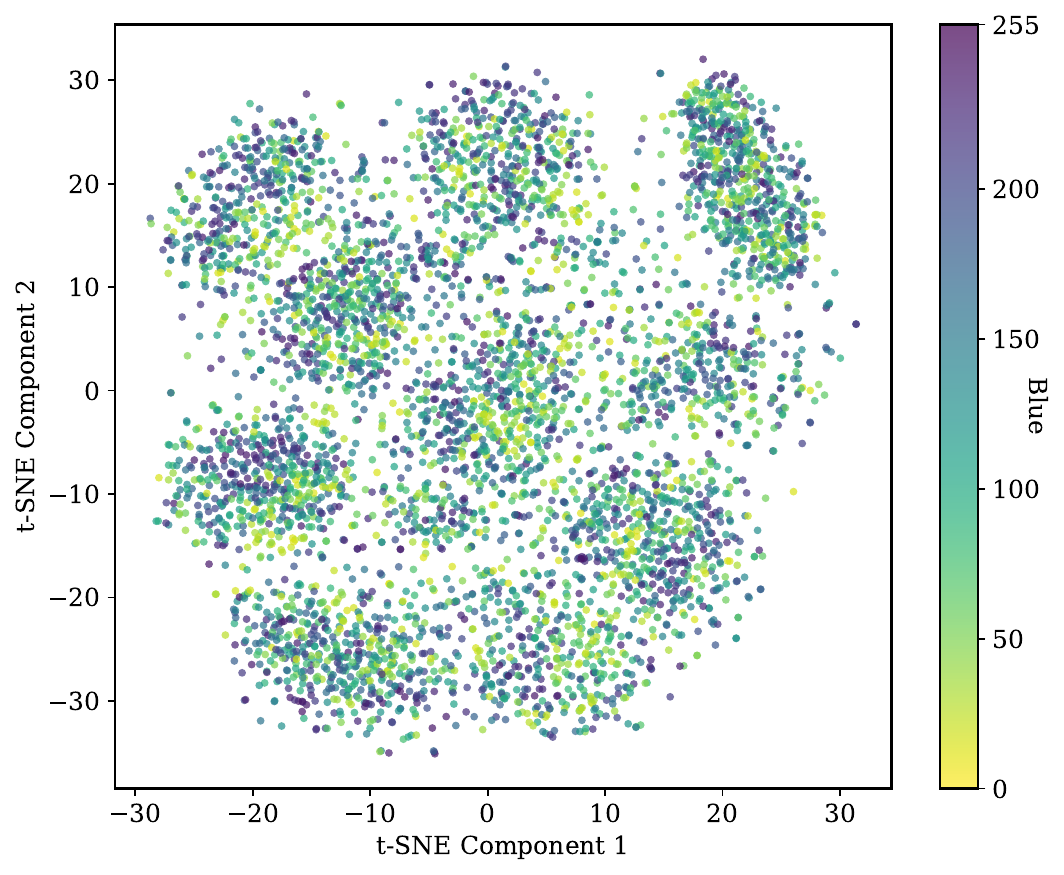}
        \caption{Unguided ($t=0$) - Blue}
        \label{fig:cmnist_tsne_uncond_blue}
    \end{subfigure}
    \begin{subfigure}{0.422\linewidth}
        \includegraphics[width=\linewidth]{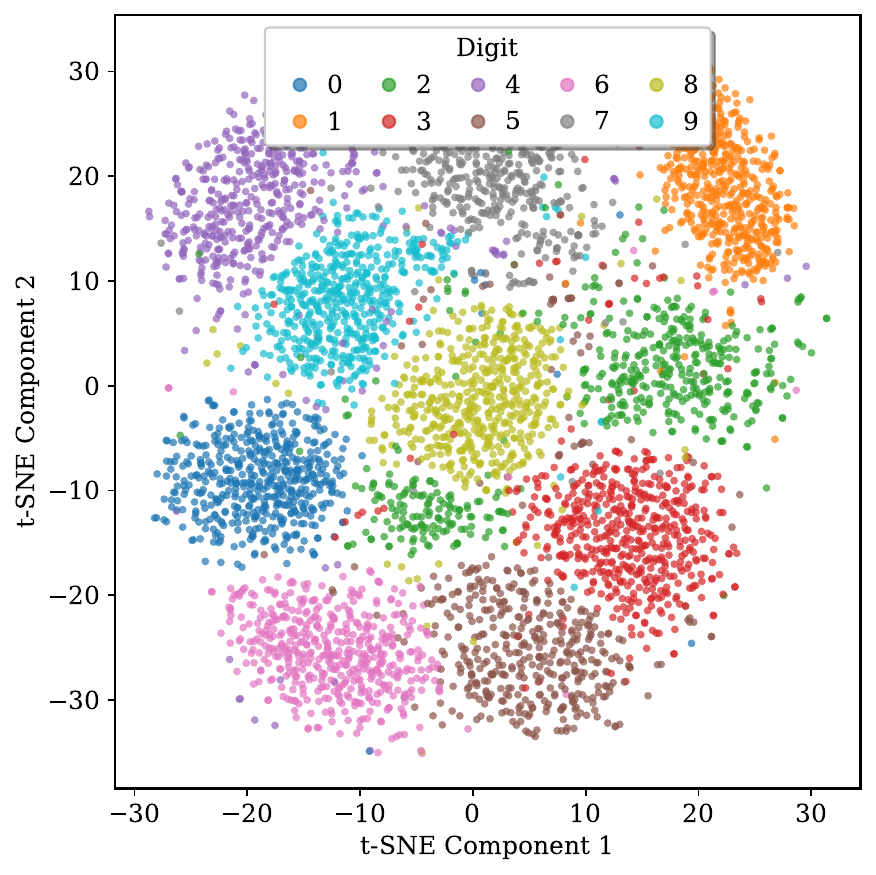}
        \caption{Unguided ($t=0$) - Class}
        \label{fig:cmnist_tsne_uncond_class}
    \end{subfigure}
    \caption{t-SNE projections of cMNIST embeddings for the unguided distributions. The reverse-guided flow in \ref{fig:cmnist_tsne_uncond_class} retains the class structure of the VAE. Figure~\ref{fig:cmnist_tsne_uncond_blue} shows how the structure of the blue color is difficult to observe in the unguided space.     \label{fig:app_mnist_unguided}}
\end{figure}

The unguided flows produced the projections shown in Figure~\ref{fig:app_mnist_unguided}.

\begin{figure}[h!]
    \centering
    \includegraphics[width=\linewidth]{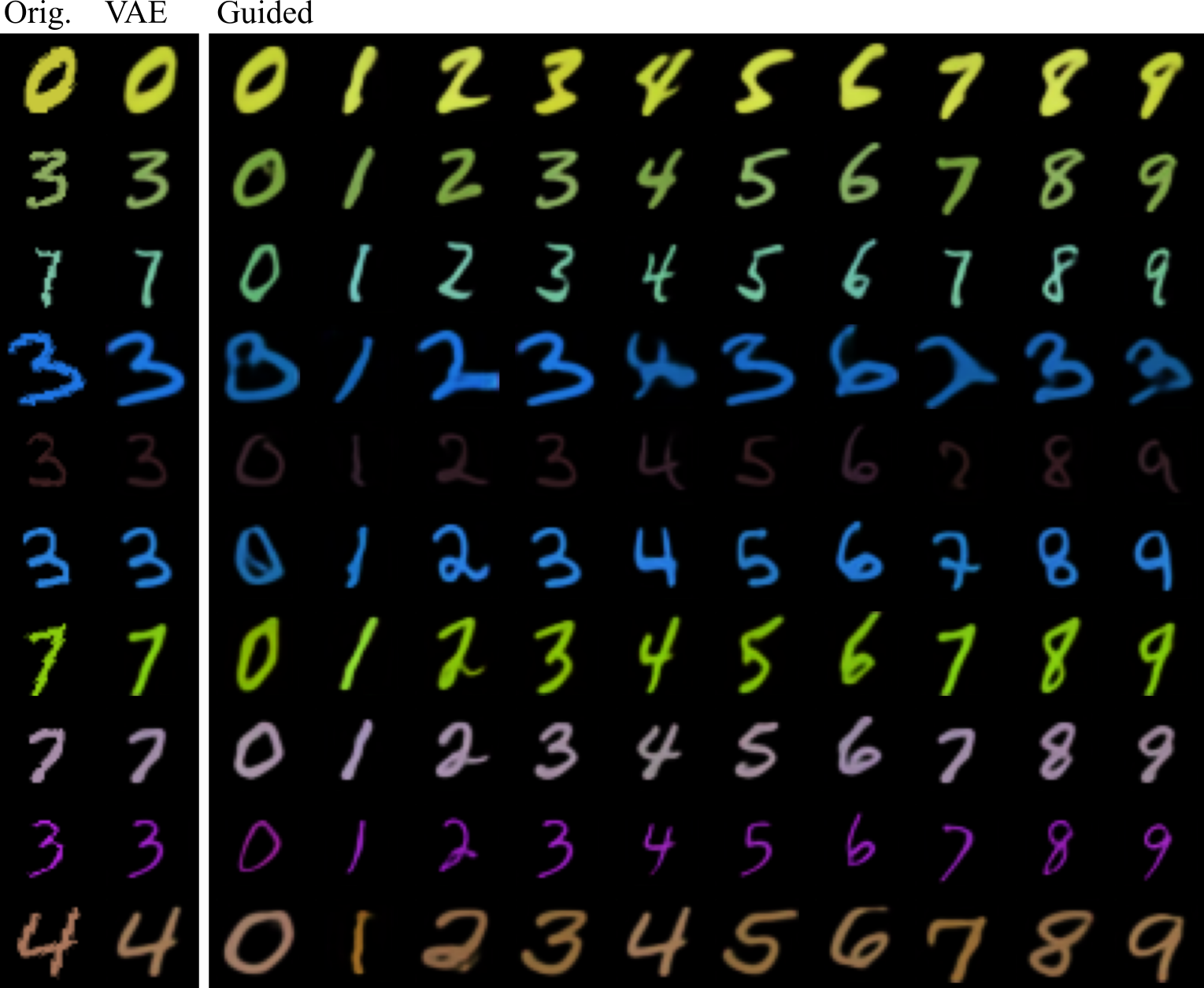}
    \caption{More examples of style transfer using the guided base representation.}
    \label{fig:guided_cfg_transfer}
\end{figure}

\begin{figure}[h!]
    \centering
    \includegraphics[width=\linewidth]{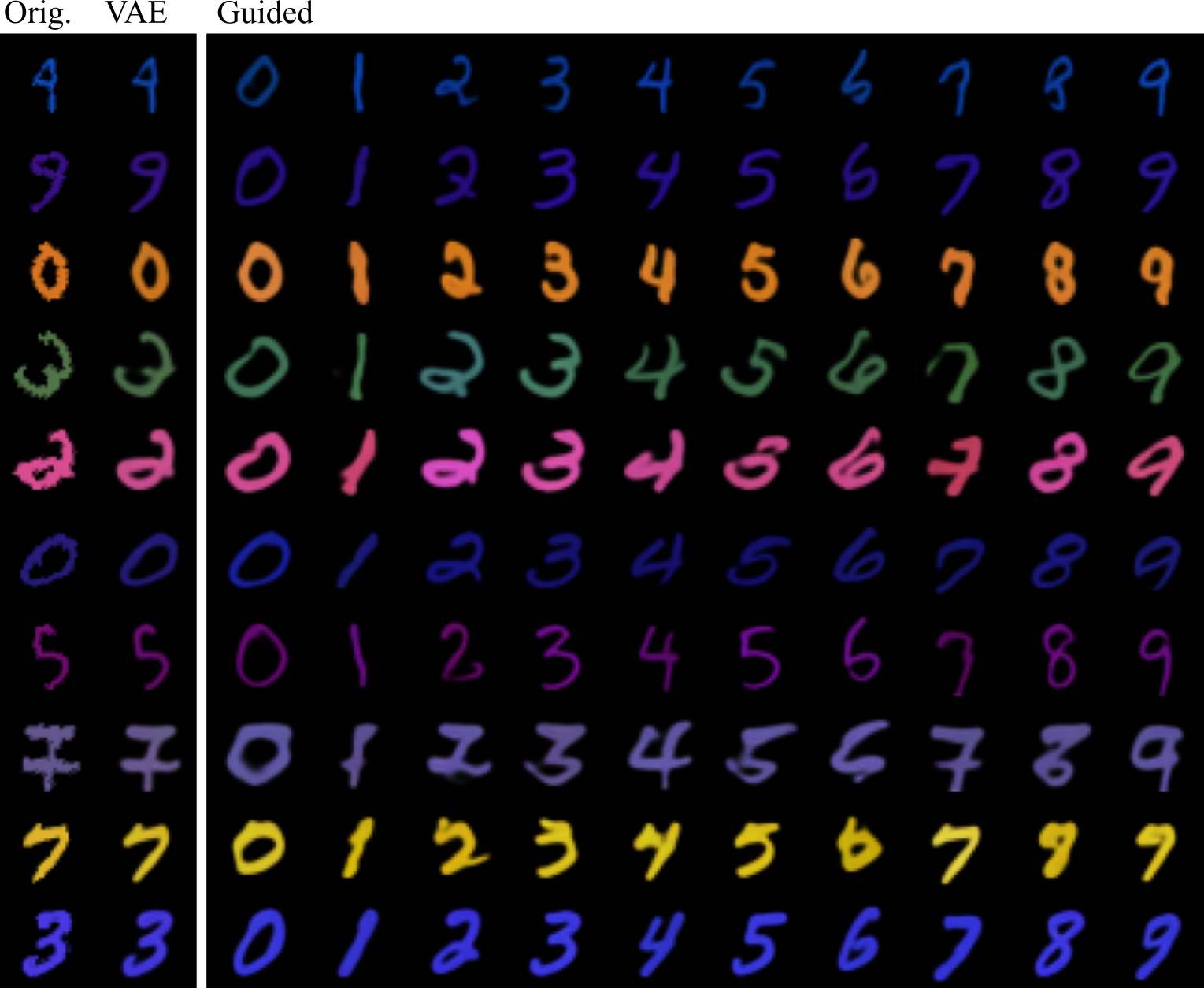}
    \caption{Examples of style transfer using the unguided base representation. Although these examples are close to the original VAE digit, they do not produce the level of correspondence seen in the guided base samples used in figure~\ref{fig:guided_cfg_transfer}.}
    \label{fig:unguided_cfg_transfer}
\end{figure}

We also consider latent seeds using the unconditional flow in Figure~\ref{fig:unguided_cfg_transfer}. These examples are close to the style of the original digit but they lack the correspondence shown using the conditional flow in Figure~\ref{fig:guided_cfg_transfer}. 

\subsection{Galaxy10 DECaLS}
\label{app:gz10}
The Galaxy10 DECaLS Dataset contains 17\,736, 256\,x\,256 sized colored galaxy images in the g, r and z band which have been scaled for clarity to RGB PNGs. These are separated in ten broad classes including: disturbed, merging, round smooth, in-between round smooth, cigar round smooth, barred spiral, unbarred tight spiral, unbarred loose spiral, edge-on without bulge and edge-on with bulge. The labels were originally provided by volunteers from the Galaxy Zoo project \citep{2011MNRAS.410..166L} and the collection was compiled by \cite{2019MNRAS.483.3255L}.

A $\beta$-VAE with $\beta=1 \times 10^{-6}$ was trained on the images to produce a $4 \times 32 \times 32$ latent representation. We used the diffusers \citep{von-platen-etal-2022-diffusers} implementation of a variational autoencoder with four down sampling blocks using the "DownEncoderBlock2D" in the encoder, each outputting 32, 64, 128 and 256 channels respectively. The decoder follows a symmetric structure using ``UpDecoderBlock2D". Each block in the VAE has four layers. This model accounts for 20.3M parameters.

We use a class conditional U-Net from the TorchCFM package \citep{2015arXiv150504597R, tong2024improving, tong2023simulation} to parametrize the velocity field using the Galaxy10 classes as the conditioning signal. The U-Net has four layers with 64, 128, 128 and 128 channels. Each down sampling of the U-Net has a single residual block. This totals 6.1M parameters. Conditioning dropout is used with a probability of $p_{cfg}=0.1$.

\begin{figure}[t]
    \centering
    \begin{subfigure}[t]{0.48\linewidth}
        \centering
        \includegraphics[width=\linewidth]{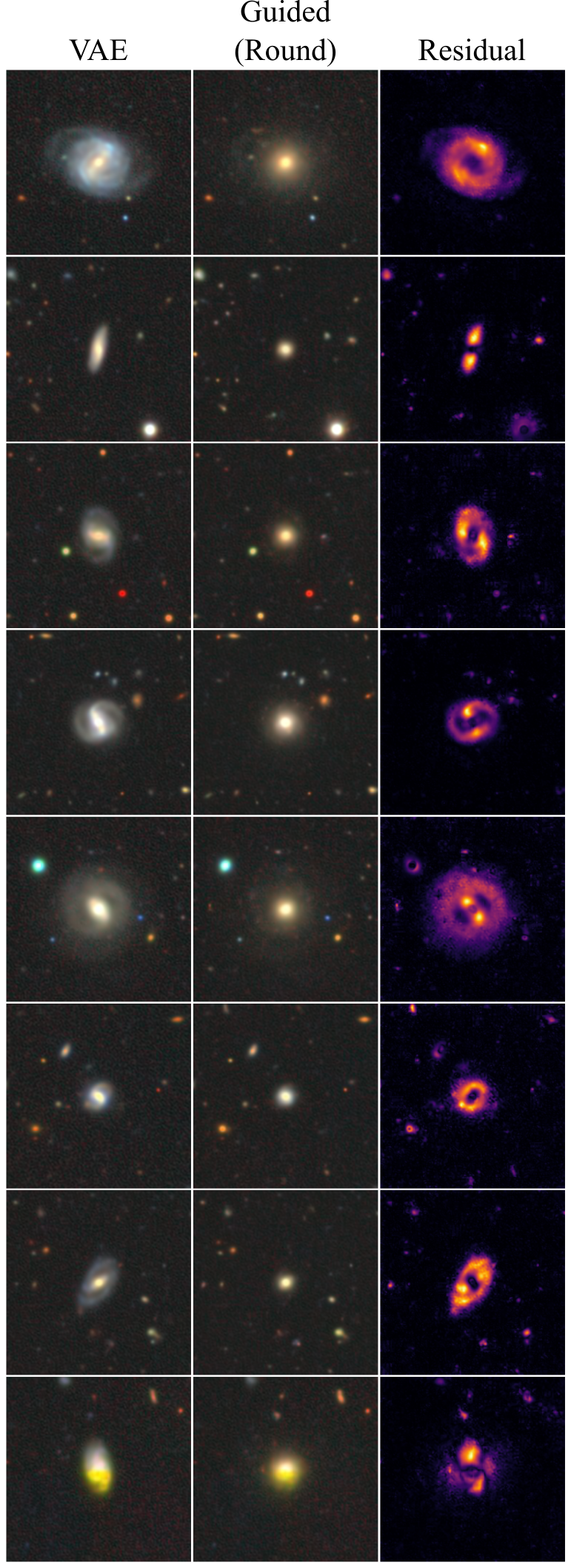}
        \caption{Barred spirals}
        \label{fig:barred_spirals}
    \end{subfigure}\hfill
    \begin{subfigure}[t]{0.48\linewidth}
        \centering
        \includegraphics[width=\linewidth]{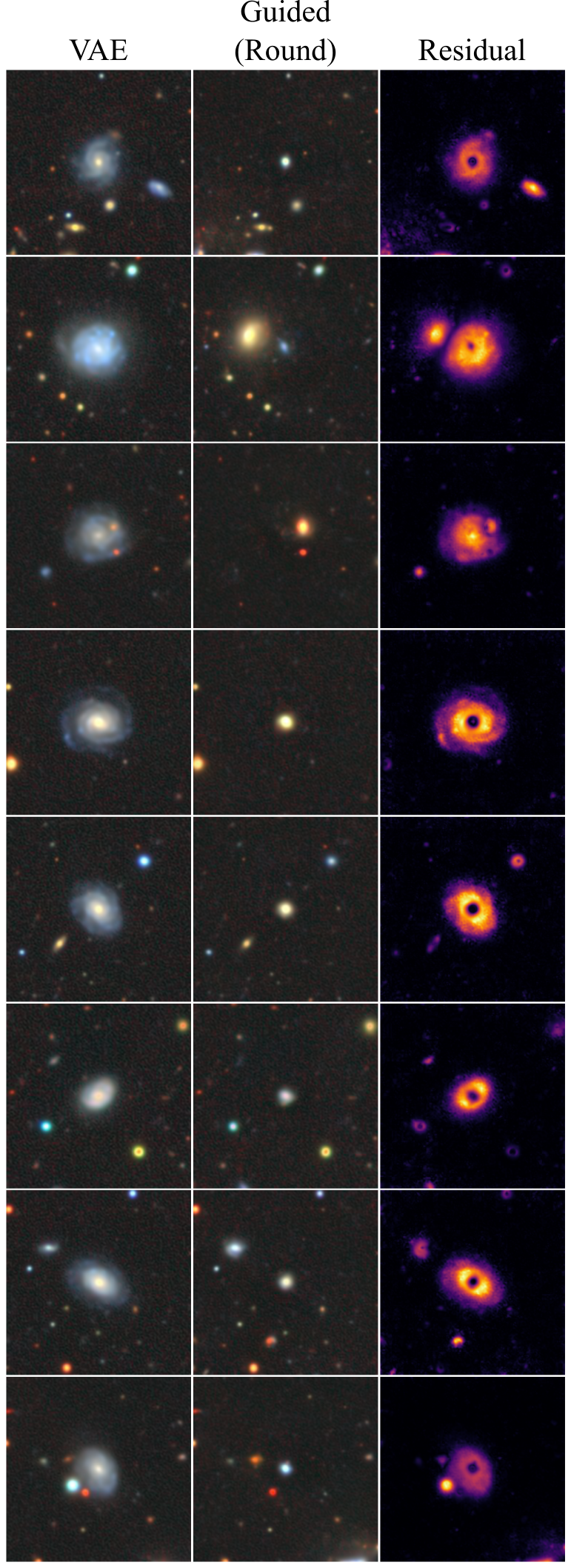}
        \caption{Tight spirals}
        \label{fig:unbarred_tight_spirals}
    \end{subfigure}
    \caption{Barred vs. unbarred tight spirals.}
    \label{fig:pair_barred_vs_tight}
\end{figure}

\begin{figure}[t]
    \centering
    \begin{subfigure}[t]{0.48\linewidth}
        \centering
        \includegraphics[width=\linewidth]{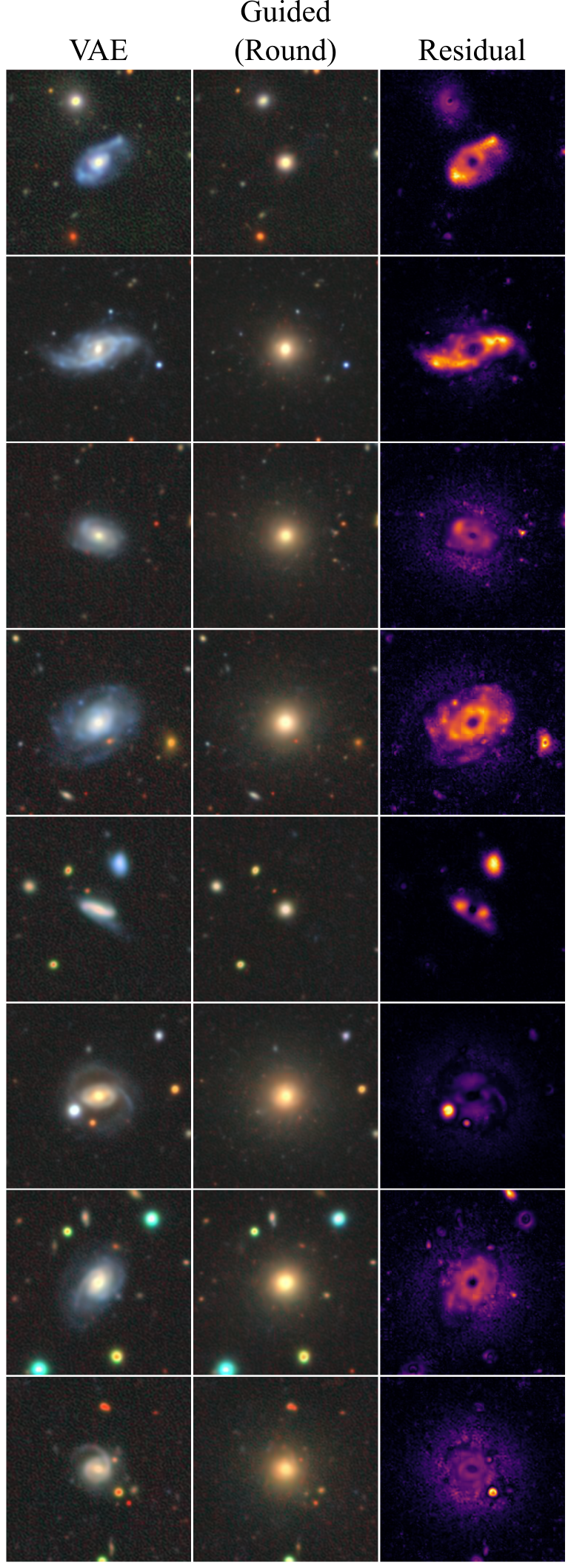}
        \caption{Unbarred loose spirals}
        \label{fig:unbarred_loose_spirals}
    \end{subfigure}\hfill
    \begin{subfigure}[t]{0.48\linewidth}
        \centering
        \includegraphics[width=\linewidth]{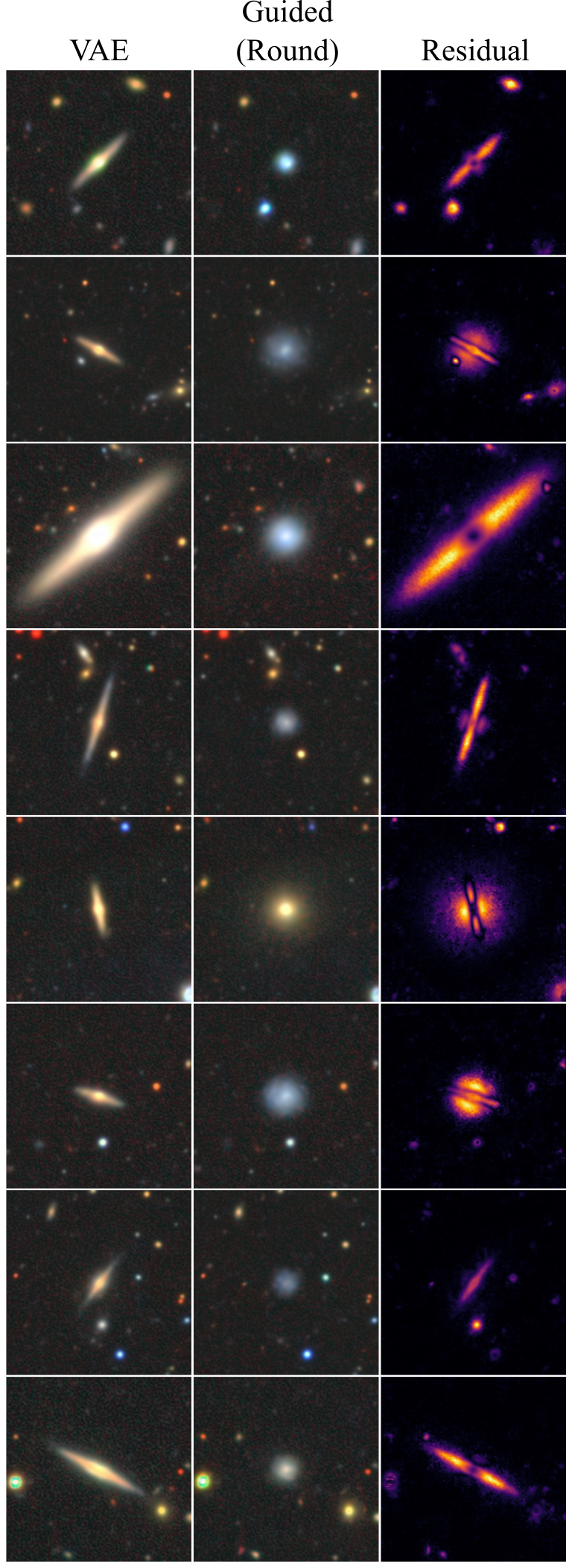}
        \caption{Edge on with bulge}
        \label{fig:additional_set}
    \end{subfigure}
    \caption{Additional feature isolation examples.}
    \label{fig:pair_loose_vs_additional}
\end{figure}

\section{Related Work}
\label{subsec: Related Work}
% Representations: PCA, Auto encoders, MAE, CLIP, BYOL, SimCLR, JEPA, AR, DLLMs, diffusion models, VLMs, CLIP, MAE, JEPA, SimCLR, BYOL
Useful low-dimensional representations of data can be recovered by various techniques. These include linear methods \citep[e.g. PCA;][]{Pearson01111901}, non-linear methods \citep[e.g. t-SNE;][]{JMLR:v9:vandermaaten08a}, and deep learning methods \citep[e.g. variational autoencoders;][]{doi:10.1126/science.1127647, kingma2022autoencodingvariationalbayes, JimenezRezende2014StochasticBA}. In addition, self-supervised learning (SSL) methods have been shown to produce useful representations for downstream tasks, including approaches such as SimCLR \citep{chen2020simple}, BYOL \citep{grill2020bootstrap}, JEPA \citep{assran2023self}, and LeJEPA \citep{balestriero2025lejepa}. Various works have investigated best practices in extracting meaningful or semantic representations from models trained using different methods across domains. This includes student teacher based systems \citep{grill2020bootstrap,oquab2023dinov2}, masked modeling \citep{He21MAE}, auto regression \citep{tao2024llms-afc}, and diffusion \citep{tang2023emergent,fuest2024diffusion,luo2023diffusion}. 

% Structured representations and limitations of unsupervised methods.
Constraining the structure of a latent space has become common practice in representation learning. This is particularly true in unsupervised disentanglement learning which hypothesizes that realistic data can be generated by a few explanatory factors of variation. For example, many approaches implicitly or explicitly penalize the total correlation between latent variables in order to encourage disentanglement \citep[]{higgins2017betavae, burgess2018understandingdisentanglingbetavae, chen2019isolatingsourcesdisentanglementvariational, jeong2019learning, shao2020controlvae, DBLP:conf/iclr/0001SB18}. However, \cite{locatello2019challenging} have shown theoretically that unsupervised learning of disentangled representations is impossible without inductive biases on the models or data. They also show empirically across many different unsupervised methods, that although the aggregated posterior of latent distributions are uncorrelated, the means of these distributions used for downstream tasks are often correlated. Additionally, unsupervised approaches typically require ground-truth factors of variation for validation \citep{higgins2017betavae, chen2019isolatingsourcesdisentanglementvariational}.

% Restrictions of transformations and a previous supervised approach.
Supervised control over input transformations has been used to create useful representations \citep{NIPS2015_ced556cd, 10.5555/3666122.3668288}. However, such approaches generally require well-defined transformations of the input data and may not be available and may not span all data features. Other works have incorporated supervised signals when learning representations. \cite{cheung2015discoveringhiddenfactorsvariation} used a cross-correlation penalty between available information and latent variables learned by an autoencoder to encourage \textit{linear} disentanglement between subspaces. This approach introduces additional terms to the reconstruction loss including a supervised and a cross-covariance term between the latent and the supervised component and therefore requires additional hyperparameters to ensure each objective is balanced and achieved during training. Finally, if new conditioning information is discovered or introduced, the full model requires re-training with an amended structure making this expensive for processes which seek to control and refine representations iteratively.

% Explicit prior and graph structure leading to expensive computational costs.
Other approaches have extended variational autoencoders to include additional conditional information \citep[e.g.,][]{kingma2014semi, NIPS2015_8d55a249, siddharth2017learningdisentangledrepresentationssemisupervised}. Such methods require the definition of priors on the conditioning information and are incorporated into the structure of the model. Additionally, the variational objective can be augmented to include (semi)-supervised losses. Training these models requires balancing the supervised loss terms with the original variational objective, requiring hyperparameter searches. This becomes even more difficult if the number of conditioning variables grow. As discussed in previous methods, if new conditioning information is discovered, the full model requires re-training with a new structure making this an inflexible approach if new or more refined conditioning information becomes available.

\section{Limitations}
This work is in early development and so it should be noted that there are a number of limitations. These include:

\begin{itemize}
    \item Solutions to ODEs used for the flow model training and inference introduce an inherent source of error. At this stage no quantification has been undertaken on how this may impact the representations during the chain, especially with the loss of information in the conditioning variables. More work is needed to understand how the capacity of the velocity field network, the simulation at inference and optimization procedures can aid or hinder representations.
    \item Computational resources were limited in the development of this paper, and therefore a full investigation into the hyper-parameters associated with the flow model training has not been undertaken. This is especially true of the latent size of the VAE and the dropout frequency used in training and how it may impact the quality of the unguided distributions.
    \item It is unclear which conditioning mechanisms are the most appropriate and efficient in approximating the guided velocity field and further work is needed to find the most effective guidance mechanisms. 
    \item Currently we only consider a Euclidean state space $\mathbb{R}^{d}$. More work is needed to understand whether this approach works on other state spaces such as discrete tokens and quantized VAEs.
\end{itemize}

\end{document}